\documentclass[acmtog]{acmart}

\AtBeginDocument{%
	\providecommand\BibTeX{{%
			\normalfont B\kern-0.5em{\scshape i\kern-0.25em b}\kern-0.8em\TeX}}}

\acmJournal{TOG}
\acmVolume{37}
\acmNumber{4}
\acmArticle{111}
\acmMonth{8}

\usepackage{color}
\usepackage{times}
\usepackage{epsfig}
\usepackage{amsmath}

\usepackage{amssymb}
\usepackage{caption}
\usepackage{multirow}
\usepackage{graphicx}
\usepackage{array}
\usepackage{booktabs}
\usepackage{subfigure}
\usepackage{soul}

\usepackage{soul, color}

\definecolor{s}{cmyk}{0.4314,0.1685,0.0441,0}
\definecolor{d}{cmyk}{0.0541, 0.4001, 0.4808, 0.0}
\definecolor{ma}{cmyk}{0.3579, 0.4895, 0, 0}
\definecolor{mi}{cmyk}{0.4135,0.0587,0.5457,0}
\definecolor{in}{cmyk}{0.0,0.0,0.0,0.349}
\definecolor{out}{cmyk}{0.0358,0.1911,0.657,0.0}
\newcommand{\sconv}{{\color{s} $\blacksquare $}}
\newcommand{\dconv}{{\color{d} $\blacksquare $}}
\newcommand{\maxpl}{{\color{ma} $\blacksquare $}}
\newcommand{\mixpl}{{\color{mi} $\blacksquare $}}
\newcommand{\inbox}{{\color{in} $\blacksquare $}}
\newcommand{\outbox}{{\color{out} $\blacksquare $}}

\definecolor{Black}{cmyk}{1.0,1.0,1.0,0}
\definecolor{Purple}{cmyk}{0.45,0.86,0,0}
\definecolor{Rosolic}{cmyk}{0.00,1.00,0.50,0}
\definecolor{Orange}{cmyk}{0,0.52,0.80,0}

\newcommand{\sbranch}{static branch}
\newcommand{\dbranch}{dynamic branch}

\acmSubmissionID{xxxx}

\begin{document}
	
	\title{SketchGNN: Semantic Sketch Segmentation with Graph Neural Networks}
	
	
	\author{Lumin Yang}
	\affiliation{%
		\institution{State Key Lab of CAD\&CG, Zhejiang University}
		\streetaddress{388 Yuhangtang Rd}
		\city{Hangzhou}
	}
	\email{luminy@zju.edu.cn}
	
	\author{Jiajie Zhuang}
	\affiliation{
		\institution{State Key Lab of CAD\&CG, Zhejiang University}
		\streetaddress{388 Yuhangtang Rd}
		\city{Hangzhou}
	}
	\email{zboom1997@gmail.com}
	
	\author{Hongbo Fu}
	\affiliation{%
		\institution{{School of Creative Media, } City University of Hong Kong}
		\city{Hong Kong}
	}
	\email{hongbofu@cityu.edu.hk}
	
	\author{Xiangzhi Wei}
	\affiliation{%
		\institution{Shanghai Jiao Tong University}
		\city{Shanghai}
	}
	
	\author{Kun Zhou}
	\affiliation{%
		\institution{State Key Lab of CAD\&CG, Zhejiang University}
		\streetaddress{388 Yuhangtang Rd}
		\city{Hangzhou}
	}
	\email{kunzhou@acm.org}
	
	\author{Youyi Zheng}
	\authornote{Corresponding author.}
	\affiliation{%
		\institution{State Key Lab of CAD\&CG, Zhejiang University}
		\streetaddress{388 Yuhangtang Rd}
		\city{Hangzhou}
	}
	\email{youyizheng@zju.edu.cn}
	
	\renewcommand{\shortauthors}{Yang, et al.}
	
	\begin{abstract}
		We introduce \emph{SketchGNN}, a {convolutional} graph neural network for semantic segmentation and labeling of freehand vector sketches. 
		We treat an input stroke-based sketch as a graph, with nodes representing the sampled points along input strokes and edges encoding the stroke structure information. To predict the per-node labels, our \emph{SketchGNN} uses graph convolution and a static-dynamic branching network architecture to 
		extract the features at three levels, i.e., point-level, stroke-level, and sketch-level. \emph{SketchGNN} significantly improves the accuracy of the state-of-the-art methods for semantic sketch segmentation (by 11.2\% in the pixel-based metric and 18.2\% in the component-based metric over a large-scale challenging SPG dataset) and has magnitudes fewer parameters than both image-based and sequence-based methods.
	\end{abstract}

	

	\maketitle
	\section{Introduction}
	Freehand sketching is becoming one of the common interaction {means} between humans and machines with the continuous iteration of digital touch devices (e.g., smartphones, tablets) and various {sketch-based interfaces on them.}
	However, sketch interpretation still remains difficult for computers due to the inherent ambiguity and sparsity in user sketches, {since sketches are often} created with varying abstraction levels, artistic forms, and drawing styles. While many previous works
	attempt to interpret a whole sketch \cite{eitz2012hdhso,eitz2012sbsr,Xu2013Sketch2Scene,Sangkloy:2016:SDL,2020SketchR2CNN}, part-level sketch analysis is increasingly required in multiple sketch applications \cite{Sarvadevabhatla:2017:SketchParse,Qi:2015:Im2Sketch,Song:2018:LSSCC,sketch2design2013,Lei2016Model}. 
	In this article, we focus on semantic segmentation and labeling of sketched objects, an essential task in finer-level sketch analysis.

	Lately, with the capacity of modern network architectures, deep-learning based sketch segmentation methods \cite{Sarvadevabhatla:2017:SketchParse,Li:2019:Fast,zhu20192d,Li:2019:TDUSPG,Qi:2019:SSegP,Wu:2018:SSeg} have greatly improved the performance over traditional sketch segmentation methods \cite{Delaye2015A,Gennari2005Combining,Schneider:2016:ESS}. These learning-based methods can be divided into two groups: \emph{image-based methods} \cite{Sarvadevabhatla:2017:SketchParse,Li:2019:Fast,zhu20192d} and \emph{sequence-based methods} \cite{Li:2019:TDUSPG,Qi:2019:SSegP,Wu:2018:SSeg}. The {image-based methods} treat {a}  sketch as a {raster}  image {and thus} unavoidably ignore the stroke structure. In contrast, the sequence-based methods use relative coordinates of stroke points and pen actions to encode the stroke structure but neglect the proximity of points (especially among different strokes). The proximity information is crucial for sketch analysis according to the Gestalt laws \cite{Wertheimer:1938:Laws}).
	
	\begin{figure}[]
		\begin{tabular}{ll}
			& \multirow{4}{*}{\includegraphics[scale=0.28]{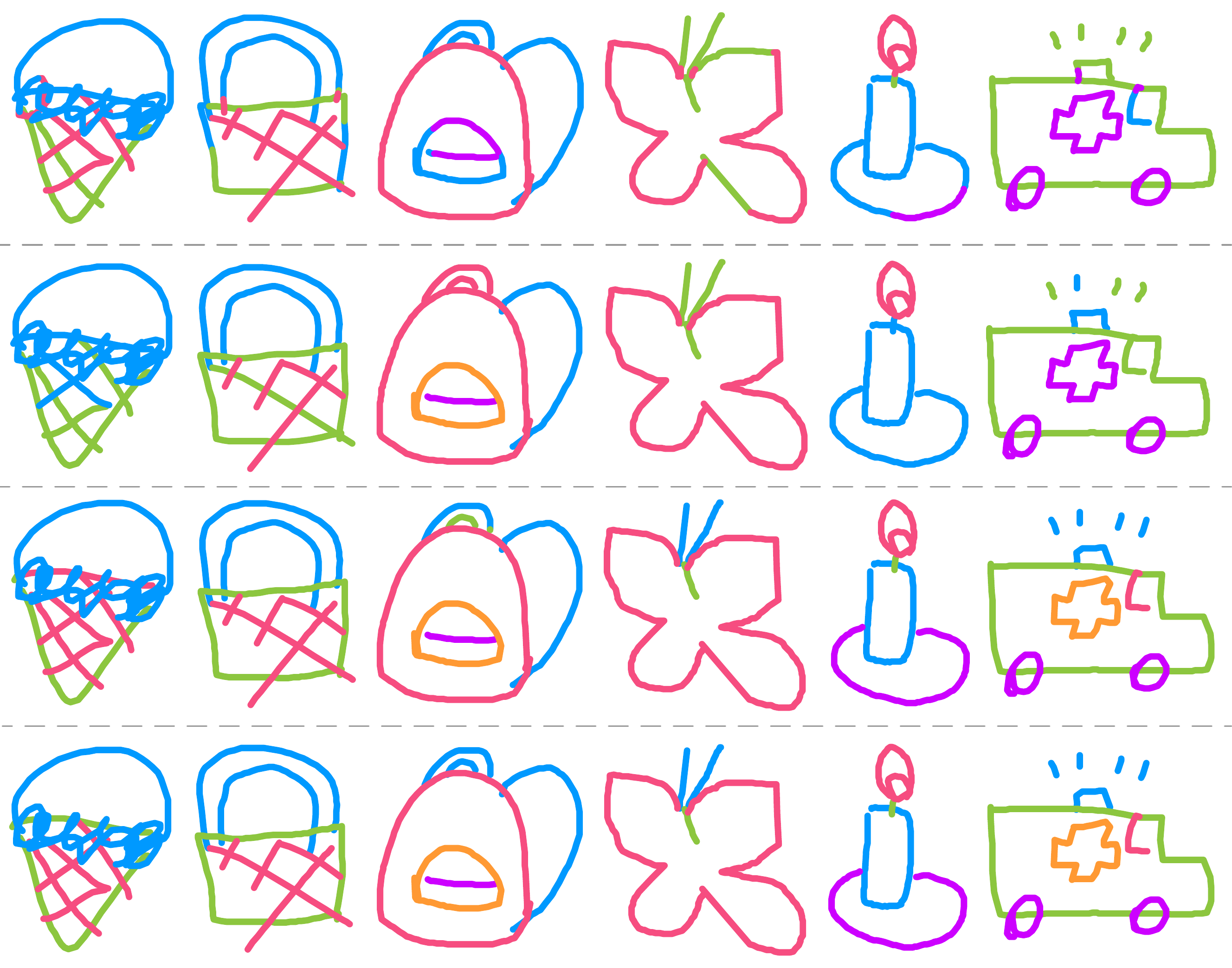}} \\
			\specialrule{0em}{-3pt}{-5pt}
			\rotatebox{90}{SPGSeg} \rotatebox{90}{\scriptsize{\cite{Li:2019:TDUSPG}}} & \\
			\rotatebox{90}{FastSeg} \rotatebox{90}{\scriptsize{\cite{Li:2019:Fast}}}& \\
			\specialrule{0em}{3pt}{10pt}
			\rotatebox{90}{Ours}  &                   \\
			\specialrule{0em}{5pt}{10pt}
			\rotatebox{90}{Human}    &                 \\
		\end{tabular}
		\caption{Our graph-based method outperforms the state-of-art image-based metheod (FastSeg) and sequence-based method (SPGSeg), achieving a similar result to the manual labeling on the SPG Dataset.}
		\label{fig:teaser}
	\end{figure}
	
	To address the above issues with the existing solutions, we adopt the graph representation \cite{GNN2009,2016Convolutional} to the sketch domain and present a novel method based on graph {neural} networks {(GNNs)}. We treat a sketch as a 2D point set with certain graphical relationships automatically built from the original stroke structure. Our graph-based representation provides richer information against a raster image representation. {Unlike sequence-based methods based on relative coordinates, our method uses} the absolute coordinates of points, thus naturally providing the proximity. Furthermore, we introduce a novel \textit{Stroke Pooling} operation to allow our network to aggregate stroke-level features in addition to point-level and sketch-level features, and greatly improve the consistency of labels within individual strokes.
	
\begin{figure*}
		\begin{center}		
			\includegraphics[width=\linewidth]{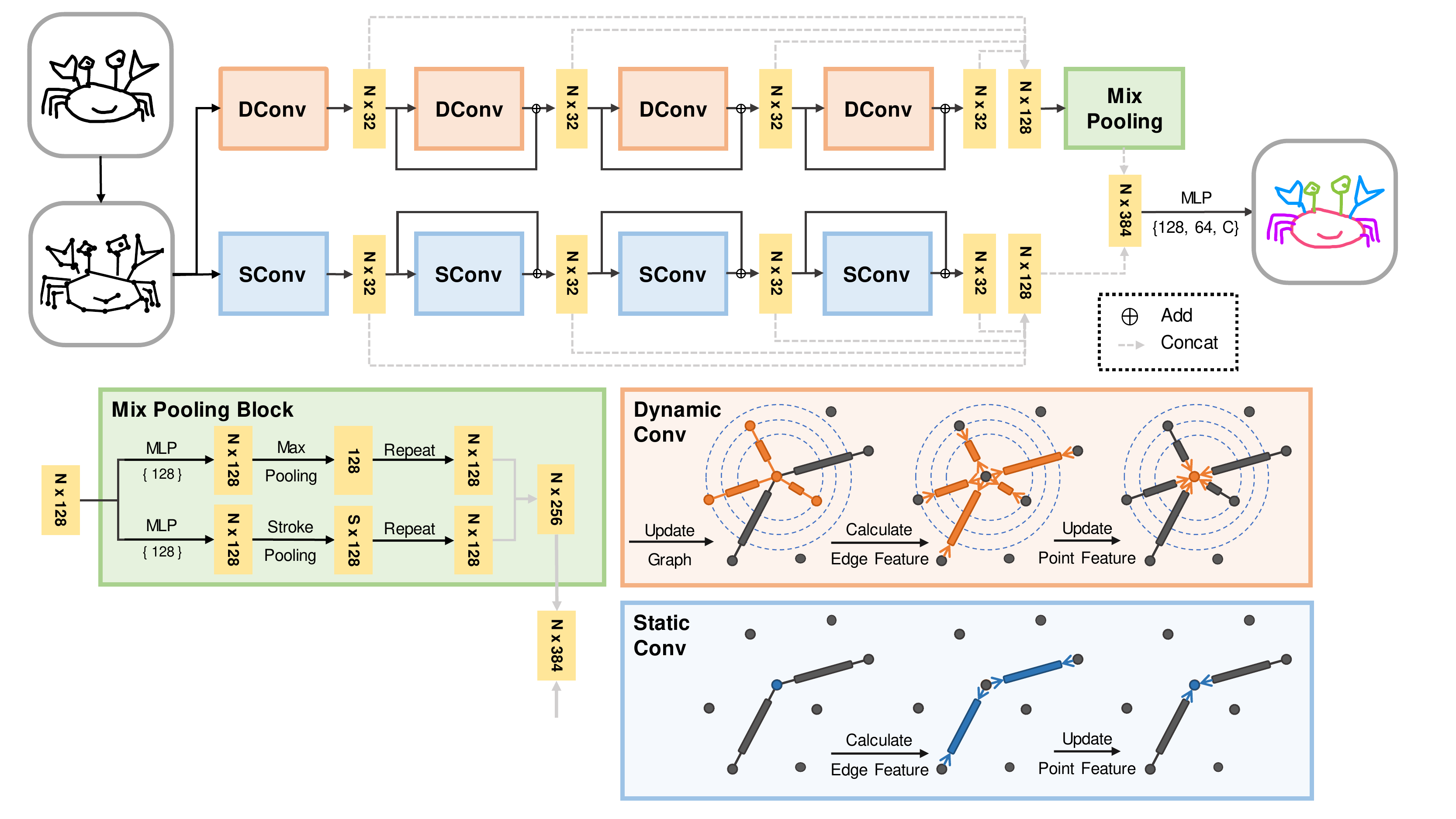}
		\end{center}
		\caption{The architecture of our SketchGNN. An input sketch is first converted into a graph {based on the stroke structure} {(the graph is simplified for illustration purpose)}. The graph node features and the connectivities are fed into the network, passing through two graph convolutional branches to extract the inter-stroke features (top, the global branch) and the intra-stroke features (bottom, the local branch). The extracted inter-stroke features are further fed into a mix pooling block to extract the global features, {which are}  subsequently concatenated with the local features, and fed into a Multi-Layer Perceptron (MLP) to get the final results.}
		\label{fig:pipeline}
	\end{figure*}

	Although the {GNN}-based method has the advantages compared to the other methods, semantic interpretation of sketches from the graphs built using the basic stroke structure (i.e., the static edges within individual strokes) is still challenging because of the resulting sparse graph structure. Schneider et al. \shortcite{Schneider:2016:ESS} manually add relation edges {(e.g., proximity and enclosing relations)} in the graph before using a Conditional Random Field (CRF). Their method, however is sensitive to variations of input. Inspired by the dynamic edges used in {3D} point cloud analysis \cite{Martin:2017:DEDF,Valsesia:2019:LLGM,Wang:2019:DGCNN}, we use a similar technique in our network to extend the basic structure information. To alleviate the problem that dynamic edges may possibly bring wrong relationships between vertices and thus contaminate the original correct structure, we propose a two-branch network (Fig. \ref{fig:pipeline}): one branch using the original sparse structure and the other with dynamic edges, to balance the correctness and sufficiency.
	
	Our main contributions are as follows: (1) We propose the first {GNN}-based method for semantic segmentation {and labeling} {of sketched objects}; (2) Our method significantly improves the accuracy of state-of-the-art and has magnitudes fewer parameters than both image-based methods and sequence-based methods.

	\section{Related work}
	\paragraph{Sketch Grouping} Sketch grouping divides strokes into clusters, {with each cluster corresponding} to an object part. Qi et al. \shortcite{Qi:2013:SPG} treat this problem as a graph partition problem, and group strokes by graph cut. Later Qi et al. \shortcite{Qi:2015:MBUOEVPG} present a grouper that utilizes multiple Gestalt principles synergistically, with a novel multi-label graph-cut algorithm. Li et al. \shortcite{Li:2018:USPG} and \shortcite{Li:2019:TDUSPG} use ordered strokes to represent a sketch, and develop a sequence-to-sequence {Variational Autoencoder} (VAE) model to learn a stroke affinity matrix. These methods, however, do not address the semantic labeling problem.
	
	\paragraph{Semantic Sketch Segmentation} To semantically segment a sketch into groups with semantic labels, early works on sketch segmentation use hand-crafted
	features with limited ability to handle the large variations of  sketches \cite{Delaye2015A,Gennari2005Combining}. 
	Such user intervention \cite{Noris2012SmartScribbles,Perteneder2015cLuster} is often needed to achieve desired segmentation results.
	Later techniques \cite{Huang:2014:DSL,Schneider:2016:ESS} leverage data-driven approaches to improve the accuracy of automatic segmentation. For example, Huang et al. \shortcite{Huang:2014:DSL} use a mixed integer programming {algorithm and utilize} 
	the segmentation information in a repository of pre-segmented 3D models. Schneider and Tyutelaars \shortcite{Schneider:2016:ESS} classify strokes based on Fisher vectors, build a graph by encoding relations between strokes, and finally use a Conditional Random Field (CRF) to solve for the most suitable label configuration. While these methods achieve reasonably accurate segmentation results, they are often computationally expensive. 
	
	The recent deep learning methods improve both the segmentation accuracy and the efficiency \cite{Sarvadevabhatla:2017:SketchParse,Li:2019:Fast,zhu20192d,Li:2019:TDUSPG,Qi:2019:SSegP,Wu:2018:SSeg}. The methods of \cite{Sarvadevabhatla:2017:SketchParse,Li:2019:Fast,zhu20192d} treat the sketch segmentation task as a semantic image segmentation problem and use convolutional neural networks (CNNs) to solve the problem. Such approaches usually ignore the structure information of  strokes or {use the stroke structure information in a post-processing step \cite{Li:2019:Fast}.} {In contrast,} the methods of \cite{Li:2019:TDUSPG,Qi:2019:SSegP,Wu:2018:SSeg} treat the task as a sequence prediction problem. They use relative coordinates and pen actions to encode the structure information. {However,} sequence-based representations ignore the proximity of points. In contrast, our method uses a graph representation to fully exploit both the stroke structure and the stroke proximity, with carefully designed convolutional operations to extract both the intra-stroke and inter-stroke features.

	
	
	\paragraph{Graph {Neural} Networks} Graph {Neural} Networks (GNNs) have been used in many applications {for example for processing social networks} \cite{Tang:2009:RLLSD}, in  recommendation engines \cite{Monti:2017:GMCRMNN,Ying:2018:GCN}, and in natural language processing \cite{Bastings:2017:GCESNMT}. {GNNs} are also suitable to process 2D and 3D point cloud data. Sketches are composed of strokes with ordered point sequences, making it possible to  construct graphs based on strokes and to use {GNNs} for sketch segmentation. As far as we know, we are the first to apply {GNNs} to semantic sketch segmentation and labeling.
	
	Graph structures in most {GNNs} are static. Recent studies about dynamic graph convolution show that changeable edges may perform better. For instance, filter weights in \cite{Martin:2017:DEDF} are dynamically generated for each specific data. \emph{EdgeConv} in \cite{Wang:2019:DGCNN} dynamically computes node neighbors and constructs new graph structures in each layer. Valsesia et al. \shortcite{Valsesia:2019:LLGM} also construct node neighbors with {the $k$-nearest neighbors (KNN) algorithm}, in order to learn to generate point clouds. Since original graphs built from {the stroke structure} are very sparse, it is difficult to learn {effective point-level} features. To better capture global and local features, we will adopt a two-branch network and use both static and dynamic graph convolution{s}. Lately, Li et al. \shortcite{Li:2019:DGCN} leverage residual connections, dense connections, and dilated convolution to solve the problem of vanishing gradient and over-smoothing in {GNNs}  \cite{Li:2018:DIGCNSL,Thomas:2017:SCGCN,gcn_survey2019}. Our method also exploits similar ideas when building our multi-layer {GNNs}.

	\section{Overview}\label{sec:overview}
	Fig.~\ref{fig:pipeline} shows the pipeline of our network. {Given an input sketch}, we first construct a graph from the basic stroke structure and use the {absolute} coordinate information as the features of the graph nodes {(Section \ref{sec:input_representation})}. Then the graph and the node features are fed into two branches {(Section \ref{sec:gc_units})}: a \sbranch consists of several static graph convolutional units; a \dbranch consists of dynamic graph convolutional units and a mix pooling block (Section~\ref{sec:three-level}), including a max pooling operation and a stroke pooling operation. The learned features of two branches are concatenated and fed into a Multi-Layer Perceptron (MLP) to get the final segmentation and labeling.
	
	The two-branch structure is tailored to the unique sketch structure, capturing both the intra-stroke information and the inter-stroke information. 
	In the static branch, the information only flows inside individual strokes since different strokes are not connected in the input graph. We use this branch to generate point-level features. While in the dynamic-graph branch, we add extra connections with the nodes found by a dilated KNN ($k$ nearest neighbors) function. We use a mix-pooling block, more specifically, a max-pooling and a stroke-pooling after the dynamic graph convolutional units to aggregate both the sketch-level features and the stroke-level features, respectively. Experiments have proved that the design of such three-level features, i.e., point-level, sketch-level, and stroke-level (Section~\ref{sec:three-level})  is beneficial to our task compared with traditional two-level features, i.e., point-level and sketch-level (see Section \ref{subsec:ablation}).
	
	\section{Methodology}\label{sec:method}
	In this section, we first explain our graph-based sketch
	representation as input to the network. Then we introduce the two graph convolutional units separately used in two branches.
	followed by the descriptions of the three-level features.
	
	\subsection{Input Representation}\label{sec:input_representation}
	Many existing sequence-based methods \cite{Li:2019:TDUSPG,Qi:2019:SSegP,Wu:2018:SSeg} use the relative coordinates to represent an input sketch. Instead, we use the absolute coordinates of sketch points, which are more suitable for our graph-based network structure.
	
	Specifically, we represent a single sketch as an $N$-point set
	$\mathcal{P} = {\{p_i=(x_i,y_i)\}}_{i=1,2,\cdots,{N}}$,
	where $x_i$ and $y_i$ are the 2D absolute coordinates of point $p_i$.
	A graph $\mathcal{G}$ is built using the basic stroke structure information, {leading to} a sparse graph $\mathcal{G} = (\mathcal{V}, \mathcal{E})$, where $\mathcal{V} = \mathcal{P}$ and $\mathcal{E}$ includes the edges that connect adjacent points on each single stroke.
	We use the same input for both the \sbranch~and the \dbranch of our network.
	
	\subsection{Graph Convolutional Units}\label{sec:gc_units}
	We use two types of graph convolutional units in our network: a \emph{static} graph convolutional unit (\emph{SConv} for short) in the \sbranch and a \emph{dynamic} graph convolutional unit (\emph{DConv}) in the \dbranch. Both units use the same graph convolutional operation. The main difference between them is that the \emph{SConv} unit does not update the graph connectivity during convolution in different layers while the \emph{DConv} unit updates the graph connectivity layer by layer using KNN. Both units use residual connections, since their performance is more stable than the general connection \cite{2018Representation}. We obtain the features $\mathcal{F}_{S}=\{f^{S}_i\}_{i=1,2,\cdots,{n_s}}$ after several SConv and the features $\mathcal{F}_{D}=\{f^{D}_i\}_{i=1,2,\cdots,{n_d}}$ after several DConv.
	
	\paragraph{Graph Convolution Operation} We use the same graph convolution operation as in \cite{Wang:2019:DGCNN} and briefly explain the operation here for the convenience of reading. Given a graph at the $l$-th layer
	$\mathcal{G}_l=(\mathcal{V}_l, \mathcal{E}_l, \mathcal{F}_l)$, where
	$\mathcal{V}_l$ {and $\mathcal{E}_l$ are the respective vertices and edges} in the graph $\mathcal{G}_l$, and
	$\mathcal{F}_l=\{f^l_i\}_{i=1,2,\cdots,n_l}$ is {a set of node features, each defined at a vertex} at the $l$-th layer.
	
	The node feature $f^{l}_{i}$ of the vertex $v_i$ in the $l$-th layer is updated by
	\begin{equation}
	\label{eqn:01}
	f^{l}_{i}=\max_{j:(i,j)\in \mathcal{E}_{l}} h_{\Theta_l}(f^{l-1}_i,f^{l-1}_j),
	\end{equation}
	where $\Theta$ is the learnable weights of the feature update operation $h_{\Theta}(\cdot)$,
	and the operation is defined as
	\begin{equation}
	\label{eqn:02}
	h_{\Theta}(f_i, f_j)=\mathrm{ReLU}(\mathrm{MLP}_\Theta(concat(f_i, f_j-f_i))).
	\end{equation}
	
	\paragraph{Graph Updating Strategy} {The \emph{SConv} units} only use the
	input graph and do not update the graph structure in different layers. In other words, we have $\mathcal{E}^{S}_{1}=\mathcal{E}^{S}_{2}=\cdots=\mathcal{E}^{S}_{l}=\mathcal{E}$ in the \sbranch.
	While the \dbranch dynamically changes the graph by adding a different edge set
	$\mathcal{E}^{dyn}_{l}$ to the input graph in different layers, leading to non-local diffusion across the strokes. More specifically, the graph used in the $l$-th layer of the \dbranch is defined as
	\begin{equation}
	\label{eqn:05}
	\mathcal{E}^{D}_l = \mathcal{E} \cup \mathcal{E}^{dyn}_{l}.
	\end{equation}
	To enlarge the receptive fields, $\mathcal{E}^{dyn}_{l}$ is designed to get dilated aggregations of the information, inspired by Li et al.~\shortcite{Li:2019:DGCN},
	\begin{equation}
	\label{eqn:03}
	\mathcal{E}^{dyn}_{l} = \{e_{ij}=\{v_i,v_j\} | v_j\in \mathcal{K}^{(d)}(v_i)\}_{i=1,\cdots,{N}}
	\end{equation}
	where $\mathcal{K}^{(d)}(v_i)$ is the $d$-dilated neighbors of vertex $v_i$. We use the same stochastic strategy at the training time as in \cite{Li:2019:DGCN}.
	
	\subsection{Three-level Features}\label{sec:three-level}
	The features provided for the final MLP layer are composed of three parts: \emph{point-level} features, \emph{stroke-level} features, and \emph{sketch-level} features. The point-level features $\mathcal{F}_{point}$ are obtained directly after several SConv units, meaning $\mathcal{F}_{point} = \mathcal{F}_{S}$.
	
	A mix-pooling block, which contains two pooling operations is designed to learn the sketch-level features $\mathcal{F}_{sketch}=\{f_i^{sk}\}_{i=1,2,...,N}$ and the stroke-level features $\mathcal{F}_{stroke}=\{f_i^{st}\}_{i=1,2,...,N}$.
	Before applying two pooling operations, we transform the features $\mathcal{F}_{D}$ by using different multi-layer perceptrons with learnable weights $\Theta_{sk}$ and $\Theta_{st}$ separately.

	We use the max pooling operation to aggregate the sketch-level features,
	\begin{equation}
	\label{eqn:06}
	f_{sk}=\max_{f_i \in \mathcal{F}_{D}} \mathrm{MLP}_{\Theta_{sk}}(f_i)
	\end{equation}
	and assign $f_{sk}$ for every point, i.e., $f^{sk}_i = f_{sk}$,
	similar to many existing methods used in 3D point cloud analysis (e.g., \cite{Charles2017PointNet,Martin:2017:DEDF}).
	
	To compute the stroke-level features, we propose a novel pooling operation, named \textit{stroke pooling}, to aggregate the features on every single stroke, 
	\begin{equation}
	\label{eqn:08}
	f^{st}_r=\max_{j \in \mathcal{V}_{\mathcal{S}_r}, f_j \in \mathcal{F}_{D}}\mathrm{MLP}_{\Theta_{st}}(f_j),
	\end{equation}
	where $\mathcal{S}_r$ represents the $r$-th stroke in a sketch, and $\mathcal{V}_{\mathcal{S}_r}$ represents the vertex set of the stroke $\mathcal{S}_r$. 
	The points in the same stroke gain the same stroke-level features $f^{stroke}_i=f^{st}_r, i\in \mathcal{V}_{\mathcal{S}_r}$.
	
	Therefore, the whole features used in the final MLP layers are a concatenation of the output of the dynamic branch (i.e., stroke-level feature $\mathcal{F}_{stroke}$ and sketch-level feature $\mathcal{F}_{sketch}$) and the output of the \sbranch (i.e., point-leval features $\mathcal{F}_{point}$):
	\begin{equation}
	\label{eqn:features}
	\mathcal{F}=concat(\mathcal{F}_{point},\mathcal{F}_{stroke},\mathcal{F}_{sketch}).
	\end{equation}
	\begin{figure*}[t!]
		\begin{tabular}{p{9mm}@{}c@{}c@{}c@{}c@{}}
			& \ \ \ \ \ \ \ \small{\textbf{SPGSeg} \cite{Li:2019:TDUSPG}}
			& \ \ \ \ \ \small{\textbf{FastSeg+GC} \cite{Li:2019:Fast}}
			& \ \ \ \ \ \ \ \  \ \ \ \ \ \ \ \ \ \ \small{\textbf{Ours}}
			& \small{\textbf{Human}} \ \ \ \ \ \ \\
			& \multicolumn{4}{c}{\multirow{5}{*}{\includegraphics[scale=0.25]{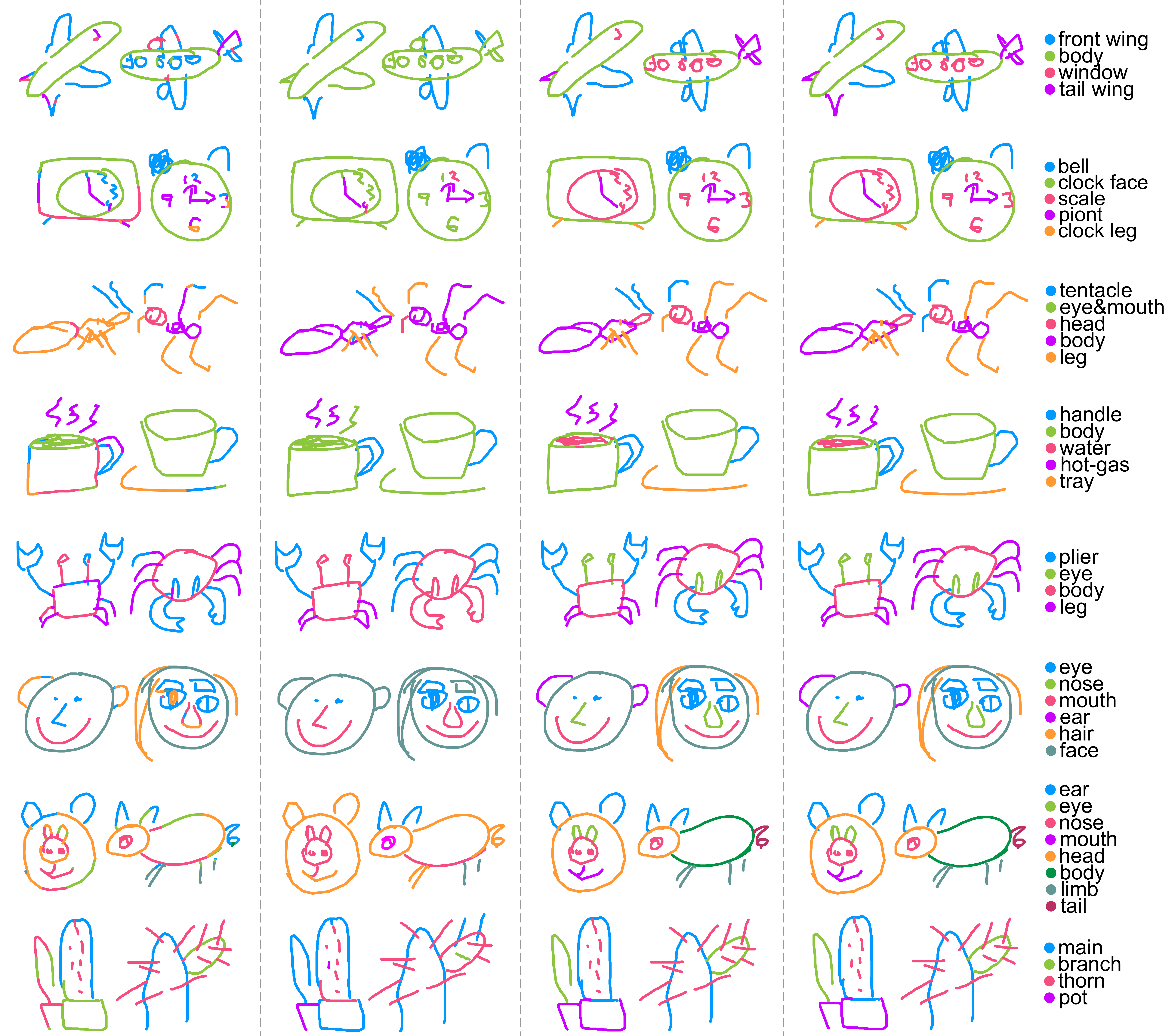}}} \\
			\specialrule{0em}{4pt}{4pt}
			airplane & \multicolumn{4}{c}{}                  \\
			\specialrule{0em}{20pt}{20pt}
			alarmclock & \multicolumn{4}{c}{}                \\
			\specialrule{0em}{19pt}{19pt}
			ant      & \multicolumn{4}{c}{}                  \\
			\specialrule{0em}{19pt}{19pt}
			coffeecup & \multicolumn{4}{c}{}                 \\
			\specialrule{0em}{19pt}{19pt}
			crab     & \multicolumn{4}{c}{}                  \\
			\specialrule{0em}{19pt}{19pt}
			face    & \multicolumn{4}{c}{}                  \\
			\specialrule{0em}{19pt}{19pt}
			pig     & \multicolumn{4}{c}{}                  \\
			\specialrule{0em}{19pt}{19pt}
			cactus   & \multicolumn{4}{c}{}                  \\
			\specialrule{0em}{15pt}{15pt}
		\end{tabular}
		\caption{Qualitative results of semantic segmentation on the SPG dataset. 
			More visual results can be found in the appendix. }
		\label{fig:res}
	\end{figure*}
	
	\subsection{Implementation Details}\label{sec:implementation_detail}
	\paragraph{Datasets} 
	We run our SketchGNN on the following four existing sketch datasets: SPG dataset~\cite{Li:2019:TDUSPG}, SketchSeg-150K dataset~\cite{Wu:2018:SSeg}, Huang14 dataset~\cite{Huang:2014:DSL} and TU-Berlin dataset \cite{eitz2012hdhso}. {The SPG and SketchSeg-150K datasets  are both built upon QuickDraw \cite{Ha2017A}, which is a vector drawing dataset selected from an online game where the players are required to draw objects in less than 20 seconds. SPG has 25 categories and 800 sketches per category and we use the same 20 categories as in \cite{Li:2019:TDUSPG}. Compared to the SPG dataset, SketchSeg-150K is relatively simpler with fewer semantic labels per categories (2-4 labels per category). With data augmentation by a sketch generative model \cite{Ha2017A}, SketchSeg-150K has about 150,000 sketches over 20 categories.  The Huang14 dataset~\cite{Huang:2014:DSL} and TU-Berlin dataset~\cite{eitz2012hdhso} are eariler smaller datasets which consist of 30 sketches and 80 sketches per category respectively.} The specific statistics of the sketch datasets are presented in Table \ref{tbl:static}.
	
	\begin{table}[h]
		\caption{Statistics of the number of labels per category and the number of storkes per sketch. The column \emph{Sketch} shows the total number of sketches in different datasets in a form of $S \times C$, where $S$ represents the number of sketches per category and $C$ represents the number of categoies in every dataset.}
		\small
		\centering
		\begin{tabular}{|c|c|c|c|c|c|c|c|}
			\hline
			\multirow{2}{*}{Dataset} & \multirow{2}{*}{Sketch} & \multicolumn{3}{c|}{strokes per sketch} & \multicolumn{3}{c|}{labels per category} \\ \cline{3-8} 
			& & min & max & median & min & max & median   \\ \hline
			SPG     & 800 $\times$  20   & 2 & 43 & 6 & 3 & 8  & 4 \\ \hline
			150k    & 7500 $\times$  20  & 1 & 7 & 3 & 2 & 4  & 3 \\ \hline
			Huang14 & 30 $\times$  10    & 9 & 123 & 48 & 3 & 11 & 6 \\ \hline
			TUB     & 80 $\times$  5    & 3 & 70 & 16 & 2 & 6  & 4 \\ \hline
		\end{tabular}
		\label{tbl:static}
	\end{table}
	
	\paragraph{{Network}}
	As shown in Fig. \ref{fig:pipeline}, our network uses $L=4$ graph convolution units in both the local branch and the global branch. Each graph convolution unit {computes edge features from connected point pairs by first concatenating point features and then using} a multi-layer perceptron with the hidden size of $32$. Then it updates point features by aggregating neighboring edge features. In the global branch, we additionally use dynamic edges found by a dilated KNN function with the number of nearest neighbors $K=8$ and an increasing dilation rate $d=1, 4, 8, 16$ for layers $0$ to $3$, respectively. In the mix pooling block, we apply a multi-layer perceptron with the hidden size of $128$ before each pooling operation. After pooling and repeating, the global features together with the local features are fed into a multi-layer perceptron with the hidden size of $[128, 64, C]$ to get the final prediction.
	
	\paragraph{Training}
	For SPG and SketchSeg-150K datasets, we directly split the original datasets to get the corresponding training and test sets.
	For Huang14 and TU-Berlin datasets, which have limited numbers of sketches per category, we make synthetic training data with labeled 3D models. We use the labeled 3D models provided by \cite{Huang:2014:DSL} for the Huang14 dataset and by \cite{2016ASAF} for the TU-Berlin dataset. We use the similar method of \cite{Li:2019:Fast} to generate such synthetic data: 
	we first render the normal maps of the 3D models using different colors for different segmentations, and then extract edge maps to approximate the sketch images by using the Canny edge detection algorithm \cite{Canny1986A}. 
	To obtain final vector sketch data, we design a simple algorithm for extracting stroke vectors from images: this algorithm 
	randomly chooses one unprocessed pixel as the seed of a current stroke and searches its adjacent pixels to find the one having the minimum included angle with the current stroke direction as the next point. The stroke direction is initialized as the horizontal direction and updated when a new pixel is processed. Some intermediate results are shown in Fig. \ref{fig:gap}. We scale all sketches to fit a $256 \times 256$ canvas. We then simplify {each sketch} using the Ramer-Douglas-Peucker algorithm \cite{Douglas1973Algorithms} and {resample it to $N$ points}. 
	
	During the training phase, we use the cross-entropy loss and Adam ($\beta_1=0.9, \beta_2=0.999$) for optimization with the base learning rate $0.002$ and batch size $64$. For SPG and SketchSeg-150K, we train the network for 100 epochs and decay the rate by 0.5 for every 50 epochs. For Huang14 and TU-Berlin, we train the network for 30 epochs and decay the rate by 0.5 for every 10 epochs. The model is implemented with PyTorch and PyTorch Geometric, a geometric deep learning extension library for PyTorch. All models are trained with an NVIDIA GTX 1080Ti GPU. {On average} {an epoch takes 15s on the SPG dataset and it takes 7ms to process a test sketch.} We use $N=256$ for the SPG, Huang14, and TU-Berlin datasets, and $N=128$ for the SketchSeg-150K dataset, since the original sketches in the latter contain fewer points. During the test time, an input sketch is first resampled to $N$ points to pass through the network and then the predicted result is mapped back to the original sketch using a nearest-neighbor scheme.

	\begin{table*}[]
		\caption{Quantitative comparisonon the SPG dataset \cite{Li:2019:TDUSPG}. ``P\_metric'' and ``C\_metric'' stand for the pixel and component metrics, respectively. GC is short for graph cut refinement \cite{Li:2019:Fast}. ``Ours w/o gt stroke'' lists the results of our method by using the structure information automatically reconstructed from the raster images of the sketches in the dataset. }
		\renewcommand\arraystretch{0.8}
		\centering
		\small
		\setlength{\tabcolsep}{1.7mm}
		\begin{tabular}{ccc|cc|cc|cc|cc}
			\hline
			\rule{0pt}{10pt}
			\multirow{2}{*}{\textbf{Category}}  &
			\multicolumn{2}{c}{\textbf{SPGSeg} \cite{Li:2019:TDUSPG}}   & \multicolumn{2}{c}{\textbf{DeepLab} \cite{deeplab2018}} & 
			\multicolumn{2}{c}{\textbf{FastSeg+GC} \cite{Li:2019:Fast}} & \multicolumn{2}{c}{\textbf{Ours}} & \multicolumn{2}{c}{\textbf{Ours w/o gt stroke}} \\ \cline{2-11}
			\rule{0pt}{10pt}
			& \textbf{P\_metric} & \textbf{C\_metric} & \textbf{P\_metric} & \textbf{C\_metric} & \textbf{P\_metric} & \textbf{C\_metric} & \textbf{P\_metric} & \textbf{C\_metric} & \textbf{P\_metric} & \textbf{C\_metric} \\ \hline
			\rule{0pt}{10pt}
			airplane         & 82.9 & 70.9 & 70.7 & 46.2 & 85.3 & 75.2 & \textbf{96.4} & \textbf{92.3} & 91.8 & 86.7 \\
			alarm\_clock     & 84.8 & 81.0 & 82.5 & 74.3 & 84.6 & 72.3 & \textbf{98.1} & \textbf{96.0} & 94.8 & 91.0 \\
			ambulance        & 80.7 & 68.1 & 72.5 & 54.2 & 85.8 & 75.3 & \textbf{94.2} & \textbf{90.3} & 89.2 & 83.0 \\
			ant              & 66.4 & 56.6 & 61.3 & 32.1 & 68.9 & 66.4 & \textbf{94.1} & \textbf{92.4} & 85.6 & 82.3 \\
			apple            & 89.9 & 71.8 & 87.3 & 60.2 & 91.4 & 82.3 & \textbf{97.2} & \textbf{93.4} & 94.3 & 88.4 \\
			backpack         & 75.2 & 63.7 & 64.3 & 28.4 & 73.3 & 59.8 & \textbf{92.7} & \textbf{86.7} & 84.8 & 77.4 \\
			basket           & 84.8 & 83.2 & 79.5 & 69.5 & 86.6 & 82.2 & \textbf{98.2} & \textbf{97.9} & 91.2 & 88.9 \\
			butterfly        & 89.0 & 83.6 & 85.6 & 69.8 & 92.7 & 79.3 & \textbf{99.6} & \textbf{98.7} & 96.3 & 94.7 \\
			cactus           & 77.5 & 72.3 & 67.2 & 30.8 & 73.3 & 68.6 & \textbf{97.5} & \textbf{96.5} & 92.7 & 80.7 \\
			calculator       & 91.1 & 89.9 & 92.5 & 92.1 & 97.4 & 93.0 & \textbf{99.3} & \textbf{99.0} & 98.7 & 95.1 \\
			campfire         & 92.3 & 91.4 & 82.9 & 83.3 & 95.6 & 92.9 & \textbf{97.3} & \textbf{96.0} & 90.3 & 87.8 \\
			candle           & 88.3 & 71.8 & 91.5 & 76.9 & 90.8 & 80.1 & \textbf{99.1} & \textbf{98.4} & 97.7 & 91.4 \\
			coffee cup       & 92.0 & 87.2 & 86.2 & 81.8 & 90.9 & 87.0 & \textbf{99.7} & \textbf{98.6} & 97.1 & 96.6 \\
			crab             & 77.9 & 70.5 & 73.9 & 49.3 & 75.9 & 55.4 & \textbf{96.1} & \textbf{94.0} & 91.7 & 87.5 \\
			duck             & 86.9 & 75.4 & 85.9 & 76.0 & 88.9 & 75.1 & \textbf{98.0} & \textbf{96.7} & 94.1 & 88.9 \\
			face             & 88.0 & 80.1 & 87.4 & 78.4 & 88.1 & 80.4 & \textbf{98.8} & \textbf{97.5} & 95.7 & 90.3 \\
			ice cream        & 85.4 & 79.3 & 80.7 & 70.3 & 87.5 & 80.1 & \textbf{95.2} & \textbf{95.3} & 89.9 & 85.3 \\
			pig              & 81.9 & 75.4 & 82.1 & 77.9 & 81.1 & 73.9 & \textbf{98.8} & \textbf{98.0} & 95.4 & 90.8 \\
			pineapple        & 89.8 & 90.2 & 85.4 & 79.5 & 91.9 & 82.3 & \textbf{98.8} & \textbf{96.5} & 94.5 & 90.7 \\
			suitcase         & 92.7 & 90.7 & 90.2 & 90.1 & 94.8 & 86.7 & \textbf{99.5} & \textbf{97.9} & 97.1 & 94.9 \\ \hline
			\rule{0pt}{10pt}
			\textbf{Average}& 84.9 & 77.6 & 80.5 & 59.0 & 86.2 & 77.4 & \textbf{97.4} & \textbf{95.6} & 93.1 & 88.6 \\ \hline
			\\
		\end{tabular}
		\label{tbl:SPG}
	\end{table*}
	
	\begin{table*}[]
		\caption{
			{Quantitative comparison} on the SketchSeg-150K dataset \cite{Wu:2018:SSeg}.}
		\renewcommand\arraystretch{0.85}
		\centering
		\small
		\setlength{\tabcolsep}{6mm}
		\begin{tabular}{c c c|c c|c c}
			\hline
			\rule{0pt}{12pt}\multirow{2}{*}{\textbf{Category}}  & \multicolumn{2}{c}{\textbf{FastSeg+GC \cite{Li:2019:Fast}}} & \multicolumn{2}{c}{\textbf{SegNet+ \cite{Wu:2018:SSeg}}}  & \multicolumn{2}{c}{\textbf{Ours}}   \\ \cline{2-7}
			\rule{0pt}{12pt}& \textbf{P\_metric} & \textbf{C\_metric} & \textbf{P\_metric} & \textbf{C\_metric}  & \textbf{P\_metric} & \textbf{C\_metric} \\ \hline
			\rule{0pt}{11pt}
			angel      & 0.98		  & 0.96		  & 0.89          & 0.86		   & \textbf{0.99} & \textbf{0.98}\\
			bird       & 0.82		  & 0.70		  & 0.98          & 0.97		   & \textbf{0.99} & \textbf{0.99}\\
			bowtie     & \textbf{1.00} & \textbf{1.00} & 0.99          & \textbf{1.00} & \textbf{1.00} & \textbf{1.00} \\
			butterfly  & 0.98		  & 0.96		  & 0.95          & 0.95		   & \textbf{1.00} & \textbf{1.00}\\
			candle     & 0.96		  & 0.78		  & 0.95          & 0.95		   & \textbf{0.98} & \textbf{0.97}\\
			cup        & 0.91		  & 0.92		  & 0.77          & 0.74		   & \textbf{0.98} & \textbf{0.98}\\
			door       & \textbf{1.00} & \textbf{1.00} & 0.99          & 0.99		   & \textbf{1.00} & \textbf{1.00}\\
			dumbbell   & 0.98		  & 0.98		  & 0.99          & 0.99		   & \textbf{1.00} & \textbf{1.00}\\
			envelope   & \textbf{1.00} & \textbf{1.00} & \textbf{1.00} & 0.99		   & \textbf{1.00} & \textbf{1.00}\\
			face       & \textbf{0.98} & \textbf{0.95}& 0.94          & 0.91		   & 0.94 & 0.92\\
			ice        & 1.00		  & 1.00		  & 0.72          & 0.69		   & \textbf{1.00} & \textbf{1.00}\\
			lamp       & 0.78		  & 0.78		  & 0.95          & 0.94		   & \textbf{0.96} & \textbf{0.96}\\
			lighter    & 0.99		  & 0.96		  & 0.99          & 0.98		   & \textbf{1.00} & \textbf{1.00}\\
			marker     & 0.90		  & 0.80		  & 0.61          & 0.55		   & \textbf{0.97} & \textbf{0.98}\\
			mushroom   & 0.98		  & \textbf{0.94} & 0.70          & 0.66		   & \textbf{0.99} & \textbf{0.94}\\
			pear       & 0.97		  & 0.94		  & 0.99          & 0.98		   & \textbf{1.00} & \textbf{1.00}\\
			plane      & 1.00		  & 0.99		  & 0.86          & 0.85		   & \textbf{1.00} & \textbf{1.00}\\
			spoon      & 0.80		  & 0.79		  & 0.85          & 0.81		   & \textbf{0.90} & \textbf{0.90}\\
			traffic    & 0.89		  & 0.93		  & \textbf{0.96} & \textbf{0.96}  & 0.95	       & 0.95		  \\
			van        & \textbf{0.99} & \textbf{0.99}& 0.87          & 0.84		   & \textbf{0.99} & \textbf{0.99}\\ \hline
			\rule{0pt}{10pt}
			\textbf{Average} & 0.95 & 0.92 & 0.90 & 0.88 & \textbf{0.98} & \textbf{0.98} \\ \hline
			\\
		\end{tabular}
		\label{tbl:150K}
	\end{table*}

	\begin{table*}[]
		\caption{Quantitative comparison on the Huang14 dataset \cite{Huang:2014:DSL}. 
		}
		\renewcommand\arraystretch{0.6}
		\scriptsize
		\centering
		\setlength{\tabcolsep}{2mm}
		\begin{tabular}{ccc|cc|cc|cc|cc|cc}
			\hline
			\rule{0pt}{7pt}
			\multirow{2}{*}{\textbf{Category}}  & \multicolumn{2}{c}{\textbf{FastSeg} \cite{Li:2019:Fast}} & \multicolumn{2}{c}{\textbf{FastSeg+GC} \cite{Li:2019:Fast}} & \multicolumn{2}{c}{\textbf{DeepLab} \cite{deeplab2018}} & \multicolumn{2}{c}{\textbf{MIP-Auto} \cite{Huang:2014:DSL}} & \multicolumn{2}{c}{\textbf{Ours}} & \multicolumn{2}{c}{\textbf{Ours+GC}}\\ \cline{2-13}
			\rule{0pt}{7pt}
			& \textbf{P\_metric} & \textbf{C\_metric} & \multicolumn{1}{c}{\textbf{P\_metric}} & \multicolumn{1}{c}{\textbf{C\_metric}} & \textbf{P\_metric} & \textbf{C\_metric} & \textbf{P\_metric} & \textbf{C\_metric} & \multicolumn{1}{l}{\textbf{P\_metric}} & \multicolumn{1}{l}{\textbf{C\_metric}} & \multicolumn{1}{l}{\textbf{P\_metric}} & \multicolumn{1}{l}{\textbf{C\_metric}} \\ \hline
			\rule{0pt}{7pt}
			airplane    & 81.1 & 65.4 & \textbf{85.5} & \textbf{75.5} & 45.0 & 30.4 & 74.0 & 55.8 &  80.0  &  66.6  & 82.9          & 75.4         \\
			bicycle     & 82.9 & 67.9 & \textbf{85.4} & \textbf{76.7} & 64.9 & 46.0 & 72.6 & 58.3 &  82.0  &  69.1  & 83.5          & 76.0         \\
			candelabra  & 74.7 & 59.2 & 77.3          & 68.0          & 58.6 & 44.1 & 59.0 & 47.1 &  78.6  &  66.7  & \textbf{81.4} & \textbf{74.3}\\
			chair       & 70.0 & 60.5 & \textbf{73.9} & \textbf{69.3} & 56.3 & 44.5 & 52.6 & 42.4 &  76.3  &  66.8  & \textbf{76.5} & \textbf{72.2}\\
			fourleg     & 79.6 & 66.5 & \textbf{83.9} & \textbf{75.8} & 64.6 & 49.1 & 77.9 & 64.4 &  80.2  &  67.8  & 82.0          & 74.9         \\
			human       & 74.8 & 61.9 & \textbf{79.2} & \textbf{71.9} & 67.6 & 55.5 & 62.5 & 47.2 &  75.5  &  66.3  & 76.5          & 71.0         \\
			lamp        & 85.7 & 78.1 & 86.5          & 80.9          & 68.3 & 64.8 & 82.5 & 77.6 &  87.1  &  79.2  & \textbf{89.8} & \textbf{86.5}\\
			rifle       & 68.5 & 56.3 & 71.4          & 67.3          & 63.8 & 50.2 & 66.9 & 51.5 &  77.9  &  67.4  & \textbf{79.3} & \textbf{73.3}\\
			table       & 77.6 & 67.3 & 79.0          & 73.1          & 64.6 & 51.9 & 67.9 & 56.7 &  78.6  &  68.0  & \textbf{81.0} & \textbf{76.7}\\
			vase        & 81.1 & 71.9 & \textbf{83.8} & 79.3          & 73.4 & 63.6 & 63.1 & 51.8 &  78.4  &  71.0  & 80.2          & \textbf{79.6}\\ \hline
			\rule{0pt}{7pt}
			\textbf{Average}  & 77.6 & 65.5 & 80.6 & 73.8 & 62.7 & 50.0 & 67.9 & 55.3 & 79.5 & 69.2 & \textbf{81.3} & \textbf{76.0}\\ \hline
			\\
		\end{tabular}
		\label{tbl:Huang14}
	\end{table*}
	
	\begin{table*}[h!]
		\caption{Quantitative comparison on subsets of the TU-Berlin dataset \cite{eitz2012hdhso}.} 
		\renewcommand\arraystretch{0.6}
		\scriptsize
		\centering
		\setlength{\tabcolsep}{3mm}
		\begin{tabular}{ccc|cc|cc|cc}
			\hline
			\rule{0pt}{7pt}
			\multirow{2}{*}{\textbf{Category}}  & \multicolumn{2}{c}{\textbf{FastSeg} \cite{Li:2019:Fast}} & \multicolumn{2}{c}{\textbf{FastSeg+GC} \cite{Li:2019:Fast}} & \multicolumn{2}{c}{\textbf{Ours}} & \multicolumn{2}{c}{\textbf{Ours+GC}} \\ \cline{2-9}
			\rule{0pt}{7pt}
			& \textbf{P\_metric} & \textbf{C\_metric} & \textbf{P\_metric} & \textbf{C\_metric} & \textbf{P\_metric}  & \textbf{C\_metric} & \textbf{P\_metric} & \textbf{C\_metric} \\\hline
			\rule{0pt}{7pt}
			airplane   & 77.6 & 63.5 & 82.1  & 72.4	 & \textbf{88.0}  & 74.1  & 87.7 & \textbf{77.0} \\
			chair      & 91.7 & 89.5 & 95.7  & 93.5  & 95.5  & 92.5  & \textbf{95.9} & \textbf{93.7} \\
			guitar     & 78.5 & 67.0 & 81.4  & 78.8	 & 91.6  & 86.0  & \textbf{92.5} & \textbf{87.7} \\
			motorbike  & 66.0 & 47.5 & 70.8  & 61.1	 & 75.2  & 66.2  & \textbf{76.5} & \textbf{70.6} \\
			table      & 92.0 & 87.3 & 94.0  & 91.1	 & 96.1  & 92.4  & \textbf{96.2} & \textbf{92.8} \\ \hline
			\rule{0pt}{7pt}
			\textbf{Average} & 81.2 & 71.0 & 84.8 & 79.4 & 89.3 & 82.2 & \textbf{89.8} & \textbf{84.4}   \\ \hline
			\\
		\end{tabular}
		\label{tbl:TUB}
	\end{table*}

	\section{Experiments}\label{sec:experiments}
	We evaluate our method on the aforementioned datasets and also compare our method to the current state-of-the-art sketch segmentation methods \cite{deeplab2018,Wu:2018:SSeg,Li:2019:Fast,Li:2019:TDUSPG}.

	We perform both qualitative and quantitative comparisons.
	Fig. \ref{fig:teaser} and \ref{fig:res} show some representative visual comparisons between our method and the methods of \cite{Li:2019:TDUSPG} and \cite{Li:2019:Fast} on the SPG dataset. The sequence-based method \emph{SPGSeg} \cite{Li:2019:TDUSPG} uses point drawing orders and their relative coordinates but ignores the proximity among strokes, leading to unsatisfactory results (Fig. 2, the first column). The image-based method \cite{Li:2019:Fast} is not aware of the stroke structure and hence mainly relies on the local image structure, also leading to inferior results.
	
	\subsection{Quantitative Evaluation}
	In this subsection, we discuss our quantitative comparisons on different datasets.
	For quantitative evaluation, we use the same evaluation metrics as the previous works \cite{Huang:2014:DSL,Li:2019:TDUSPG,Wu:2018:SSeg,Li:2019:Fast}: 
	\begin{itemize}
		\item Pixel-based accuracy (P\_metric), which evaluates the percentage of the pixels that are correctly labeled in all the sketches. 
		The pixel metric is sensitive to label errors that appear on large components.
		\item  Component-based accuracy (C\_metric), which evaluates the percentage of the correctly labeled strokes, irrespective of the number of pixels in one stroke. A stroke is correctly labeled if at least 75\% of its pixels have the correct label. The component metric is sensitive to label errors that appear on small components.
	\end{itemize}
	
	Table \ref{tbl:SPG} lists the quantitative results of different methods on the SPG dataset. We use the same set of data split as in \cite{Li:2019:TDUSPG}. Our approach outperforms others by a large margin: on average 11.2\% higher in terms of the pixel metric and 18.2\% higher in terms of the component metric on the SPG dataset than \emph{FastSeg + GC} \cite{Li:2019:Fast}, which performs the best among the existing methods. 
	
	To validate the benefits of stroke structure, we test our model on a reconstructed SPG dataset, in which the stroke structure is derived from the rasterized images of the sketches. Specifically, we render the vector sketches to the binary images and rebuild the vector sketches by the stroke-extraction algorithm similar to the one we used in making the synthetic data for Huang14 and TU-Berlin datasets (Section \ref{sec:implementation_detail}). 
	Considering the ambiguity of the reconstructed stroke structure, we evaluate the segmentation results on the original stroke structure after mapping the predicted label to the nearest point in the original sketch.
	The results of our method on this reconstructed SPG dataset are listed in Table \ref{tbl:SPG} (the column `Ours w/o gt stroke'). 
	With the stroke structure automatically inferred from the raster sketch images,
	our model still raises 6.9\% on average in terms of the pixel metric and 11.2\% on average in terms of the component metric compared with \emph{FastSeg + GC} \cite{Li:2019:Fast}. This indicates that our method is potentially beneficial for semantic segmentation of raster sketch images.

	Table \ref{tbl:150K} shows the comparison results on the SketchSeg-150K datasets with the same data split as in \cite{Wu:2018:SSeg}. 
	Our approach exceeds \emph{FastSeg + GC} \cite{Li:2019:Fast} 3\% higher in the pixel metric and 6\% higher in the component metric on average. The less significant performance gain by our method on the SketchSeg-150K dataset than the SPG dataset is mainly because this dataset is labeled coarsely with fewer semantic labels per category (2-4 labels per category in SketchSeg-150K versus 3-7 labels per category in SPG) and thus less challenging for the existing methods.
	
\begin{figure}[]
		\centering
		\includegraphics[width=\linewidth]{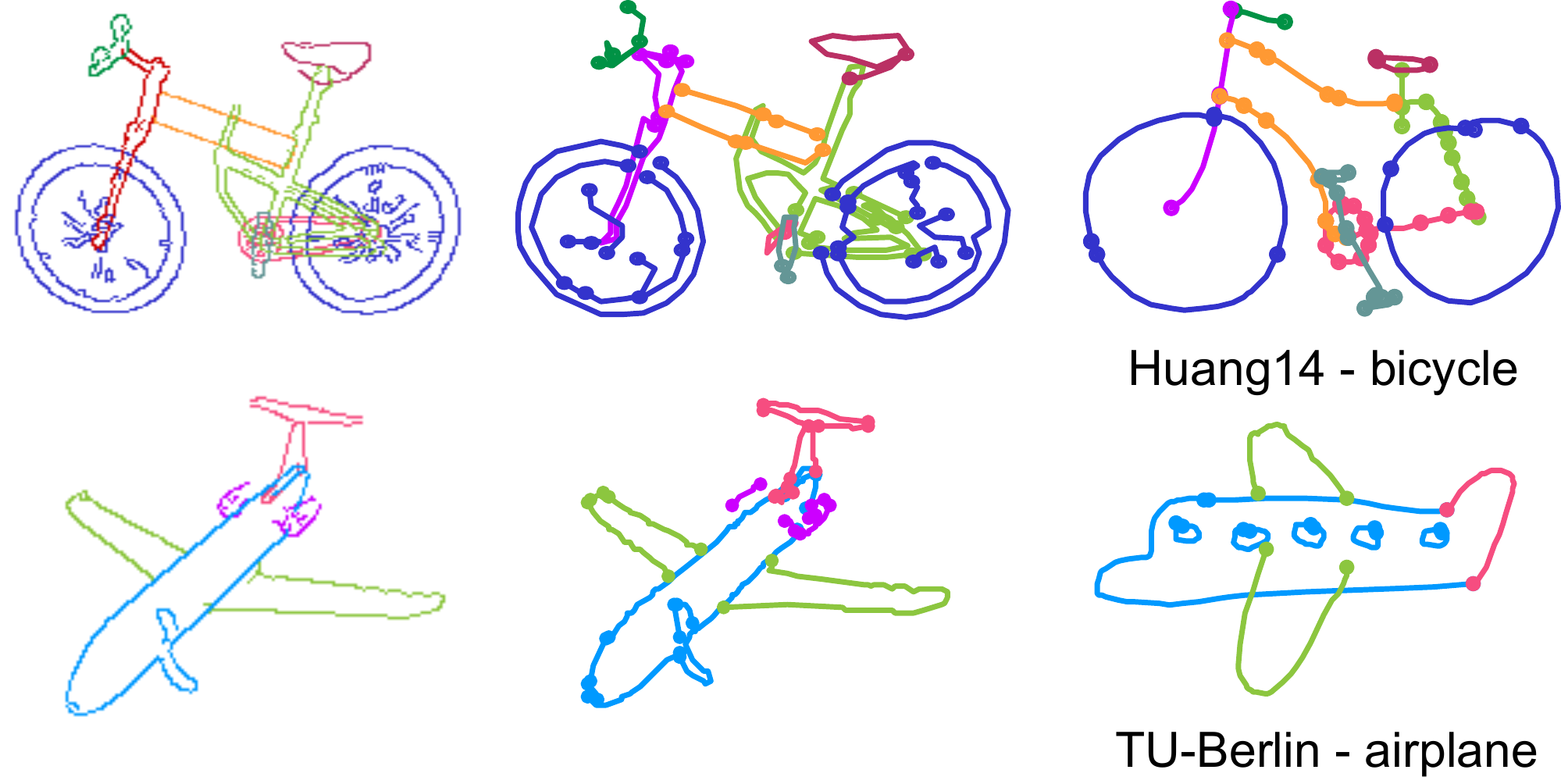}
		\caption{{To create training data for the Huang14 and the TU-Berlin datasets, we construct graphs (Middle) from edge map images (Left) rendered from 3D models.} The right column shows the exemplar freehand sketches in the test data. {The synthesized and real sketches are different in both structure and shape.} The endpoints of each stroke are marked with solid circles.}
		\label{fig:gap}
	\end{figure}

	We also evaluate our model on the Huang14 dataset and a subset of TU-Berlin dataset. Tables \ref{tbl:Huang14} and \ref{tbl:TUB} show the respective quantitative results. {In the Huang14 dataset, the individual strokes typically contain many spacious small segments (e.g., see Fig. \ref{fig:gap}, top row). Such small segments would potentially increase the structural noise and thus degrade the performance of our \emph{DConv} units, since \emph{DConv} connects new edges within the feature space. For a fair comparison}, we apply an additional graph cut algorithm to refine our results as in \cite{Li:2019:Fast}. {The situation is improved (see statistics in Table \ref{tbl:TUB}) in the TU-Berlin dataset where there are not many such spacious small segments, which indicates that our model can learn the stroke structure information. We did not {include the results of our method with graph cut} when comparing to existing methods on the SPG and SketchSeg-150K datasets, {since GC did not bring any obvious improvements.}} Note GC is more helpful in \cite{Li:2019:Fast} because their image-based method is not aware of the stroke structure and hence the prediction usually contains many spacious tiny segments within a single stroke (see details in \cite{Li:2019:Fast}). 
	
	Overall, our method {+ GC} gains on average 0.7\% higher in the pixel metric and 2.2\% higher in the component metric on the Huang14 dataset {(than {\emph{FastSeg + GC}})}, and on average 5.0\% higher in the pixel metric and 5.0\% higher in the component metric on the TU-Berlin dataset. On the Huang14 dataset, our results are {only} slightly better than {those by} the CNN-based method {\emph{FastSeg + GC}} \cite{Li:2019:Fast} (in some categories {even} slightly worse, Table \ref{tbl:Huang14}). {This is mainly due to} the large domain gap between the synthetic data {rendered from 3D models} and the real hand-drawn data, as shown in Fig. \ref{fig:gap}. The large domain gap may result in large structural noise for {GNN}-based methods to fully capture the stroke structures. Our method, however is still able to achieve the state-of-the-art performance.

	It is noteworthy that compared with existing deep network-based models for sketch segmentation, our model has orders of magnitude fewer parameters. For example, the sequence-based method SPGSeg \cite{Li:2019:TDUSPG} has a parameter size of 23.4MB while the image-based method \emph{FastSeg + GC} \cite{Li:2019:Fast} has a parameter size of 40.9MB. In contrast, the parameter size of our model is only 434KB, which is two orders of magnitude tinier than them. However, the {GNN} model has some additional computation, such as for the aggregation of the vertices. We implement our model with PyTorch Geometric and achieve a similar runtime performance to the state-of-the-art method (\emph{FastSeg + GC}\cite{Li:2019:Fast}). The need of fewer parameters means that our model has more potential to be developed in light-weight applications like those on mobile devices.

	\begin{table*}[]
		\caption{Quantitative results of ablation study on the SPG dataset. The baseline has only one dynamic branch and uses max pooling. \textbf{B + D.S.}, \textbf{B + D.D.}, and \textbf{B + S.D.} are the baseline with different combinations of two branches (\textbf{S.} and \textbf{D.} standing for the static and dynamic branches, respectively). \textbf{B + S.P.} is the baseline with stroke pooling. \textbf{B + S.D. + S.P} is our full model. We also show the simplified pipeline of each network for intuitive understanding.\sconv~and \dconv~represent the SConv and DConv blocks, respectively. \maxpl~and \mixpl~represent max-pooling and mix-pooling, respectively. \inbox \ represents the input features and \outbox \ represent the features sent into the final MLP.}
		\renewcommand\arraystretch{0.9}
		\small
		\centering
		\setlength{\tabcolsep}{2.3mm}
		\begin{tabular}{ccc|cc|cc|cc|cc|cc}
			\hline
			\rule{0pt}{10pt}
			& \multicolumn{2}{c}{\textbf{Baseline}}     
			& \multicolumn{2}{c}{\textbf{B + D.S.}} 
			& \multicolumn{2}{c}{\textbf{B + D.D.}} 
			& \multicolumn{2}{c}{\textbf{B + S.D.}} 
			& \multicolumn{2}{c}{\textbf{B + S.P.}}          
			& \multicolumn{2}{c}{\textbf{B + S.D. + S.P.}} \\
			
			\vspace{1mm}
			& \multicolumn{2}{c|}{\includegraphics[scale=0.5]{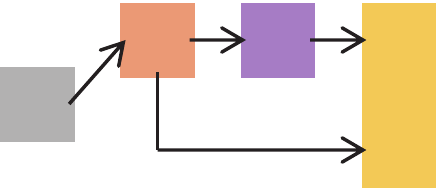}} 
			& \multicolumn{2}{c}{\includegraphics[scale=0.5]{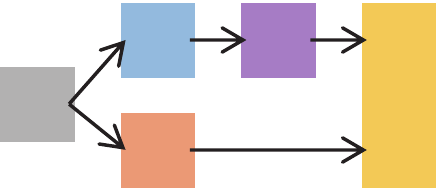}}  
			& \multicolumn{2}{c}{\includegraphics[scale=0.5]{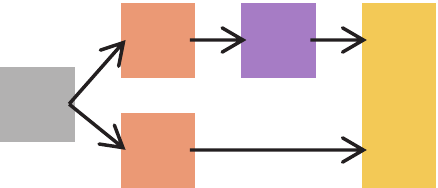}}  
			& \multicolumn{2}{c|}{\includegraphics[scale=0.5]{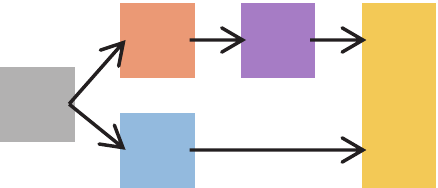}} 
			& \multicolumn{2}{c|}{\includegraphics[scale=0.5]{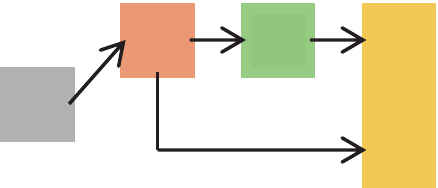}} 
			& \multicolumn{2}{c}{\includegraphics[scale=0.5]{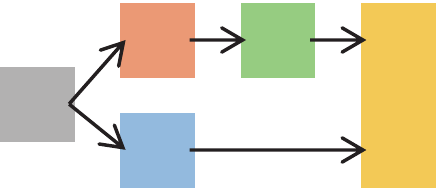}}  \\
			
			\cline{2-13}
			
			\rule{0pt}{10pt}
			\textbf{Category} & \textbf{P}  & \textbf{C} & \textbf{P}  & \textbf{C} & \textbf{P}  & \textbf{C} & \textbf{P}  & \textbf{C} & \textbf{P}  & \textbf{C}& \textbf{P}  & \textbf{C} \\ \hline
			\rule{0pt}{10pt}
			airplane      & 88.00  & 80.75 	& 81.60 & 72.19    & 88.29 & 78.32  & 91.34 & 83.07      & 95.94 & 90.67    & \textbf{96.13} & \textbf{92.00} \\   
			alarm\_clock  & 92.15  & 86.79 	& 88.69 & 81.08    & 90.74 & 84.98  & 93.82 & 89.57      & 97.57 & 95.04    & \textbf{97.99} & \textbf{96.08} \\   
			ambulance     & 82.04  & 74.49 	& 76.94 & 54.36    & 82.65 & 73.14  & 89.20 & 83.61      & 92.48 & 88.14    & \textbf{93.92} & \textbf{89.98} \\   
			ant           & 82.70  & 78.55 	& 76.80 & 66.81    & 84.35 & 78.09  & 87.93 & 83.65      & 90.30 & 89.21    & \textbf{94.12} & \textbf{92.42} \\   
			apple         & 91.49  & 78.39 	& 88.07 & 64.64    & 92.32 & 75.96  & 93.44 & 83.07      & 96.41 & 91.32    & \textbf{97.07} & \textbf{92.32} \\   
			backpack      & 77.50  & 63.19 	& 61.19 & 39.45    & 75.45 & 60.77  & 79.58 & 61.49      & \textbf{92.68} & \textbf{86.36}    & 92.20 & 85.16 \\   
			basket        & 82.12  & 76.60 	& 80.92 & 74.25    & 84.40 & 76.41  & 83.42 & 77.49      & 97.52 & 97.13    & \textbf{97.75} & \textbf{97.32} \\   
			butterfly     & 94.04  & 88.96 	& 94.19 & 90.79    & 93.16 & 89.04  & 96.47 & 91.56      & 98.76 & 96.96    & \textbf{99.61} & \textbf{98.71} \\   
			cactus        & 89.04  & 80.20 	& 88.94 & 80.25    & 88.68 & 82.08  & 92.67 & 86.43      & 95.81 & 94.16    & \textbf{96.85} & \textbf{95.89} \\   
			calculator    & 97.66  & 96.45 	& 97.33 & 96.29    & 97.46 & 96.53  & 98.13 & 97.03      & 99.08 & 98.02    & \textbf{99.18} & \textbf{98.49} \\   
			campfire      & 85.85  & 83.53 	& 85.29 & 76.02    & 86.11 & 80.77  & 88.17 & 86.30      & 95.44 & 92.94    & \textbf{96.06} & \textbf{94.21} \\   
			candle        & 96.40  & 91.52 	& 96.23 & 87.61    & 96.23 & 89.29  & 97.73 & 95.19      & 99.17 & 97.98    & \textbf{99.14} & \textbf{98.35} \\   
			coffee\_cup   & 95.13  & 95.21 	& 95.34 & 95.27    & 95.40 & 94.68  & 97.01 & 95.97      & 98.64 & 96.95    & \textbf{98.82} & \textbf{97.76} \\   
			crab          & 89.02  & 82.61 	& 86.90 & 78.97    & 88.10 & 82.75  & 92.43 & 86.67      & 95.03 & 92.43    & \textbf{96.08} & \textbf{94.01} \\   
			duck          & 92.82  & 88.22 	& 90.89 & 87.83    & 90.77 & 86.91  & 94.88 & 91.63      & 97.81 & 96.94    & \textbf{98.04} & \textbf{96.64} \\   
			face          & 93.58  & 89.15 	& 90.82 & 83.92    & 93.58 & 87.09  & 94.67 & 91.27      & 98.13 & 96.04    & \textbf{98.81} & \textbf{97.53} \\   
			ice cream     & 86.39  & 82.29 	& 84.94 & 79.52    & 85.44 & 77.04  & 89.78 & 84.16      & 94.62 & 92.55    & \textbf{95.21} & \textbf{95.32} \\   
			pig           & 91.14  & 86.76 	& 87.59 & 81.33    & 90.63 & 88.14  & 95.09 & 93.19      & 97.77 & 95.90    & \textbf{98.83} & \textbf{97.97} \\   
			pineapple     & 92.01  & 89.55 	& 88.76 & 85.02    & 92.67 & 88.63  & 96.13 & 92.00      & 98.69 & 95.80    & \textbf{99.03} & \textbf{96.45} \\   
			suitcase      & 95.74  & 94.10 	& 95.62 & 92.46    & 96.44 & 95.31  & 97.77 & 95.08      & 99.03 & 97.33    & \textbf{99.53} & \textbf{97.93} \\    \hline
			\rule{0pt}{10pt}																 														   					    
			\textbf{Average} & 89.74  & 84.37 	& 86.85 & 78.40    & 89.64 & 83.30  & 92.48 & 87.42    & 96.54 & 94.09    & \textbf{97.22} & \textbf{95.23} \\ \hline
			\\
		\end{tabular}
		\label{tbl:albation2}
	\end{table*}

	\begin{table*}[]
		\caption{Quantitative results of {GNN} variants on the SPG dataset. SAGE-mean and SAGE-max are GraphSAGE with the mean and pooling aggregators, respectively. GIN-$\epsilon$ has a learnable parameter $\epsilon$ initialized with 0.1, while GIN-0 sets $\epsilon = 0$. 
		}
		\renewcommand\arraystretch{0.85}
		\small
		\centering
		\setlength{\tabcolsep}{3mm}
		\begin{tabular}{ccc|cc|cc|cc|cc|cc|cc}
			\hline
			\rule{0pt}{10pt}
			\multirow{2}{*}{\textbf{Category}} & 
			\multicolumn{2}{c}{\textbf{EdgeConv}}  & \multicolumn{2}{c}{\textbf{MRGCN}}    & 
			\multicolumn{2}{c}{\textbf{SAGE-mean}} & \multicolumn{2}{c}{\textbf{SAGE-max}} & 
			\multicolumn{2}{c}{\textbf{GIN-0}}     & \multicolumn{2}{c}{\textbf{GIN-$\epsilon$}} &
			\multicolumn{2}{c}{\textbf{ECC}} \\ \cline{2-15}
			\rule{0pt}{10pt}
			& \textbf{P}  & \textbf{C} & \textbf{P}  & \textbf{C} & \textbf{P}  & \textbf{C} & \textbf{P}  & \textbf{C} & \textbf{P}  & \textbf{C} & \textbf{P}  & \textbf{C} & \textbf{P}  & \textbf{C}\\ \hline
			\rule{0pt}{10pt}
			airplane         & \textbf{96.6} 	& \textbf{92.5} 	& 95.1 	& 91.2 	& 93.5  & 88.3 	& 94.6 	& 89.9 	& 95.1 	& 89.8 	& 94.1 	& 88.7  & 94.2 & 87.2\\
			alarm\_clock     & \textbf{97.4} 	& \textbf{94.6} 	& \textbf{97.4} 	& 94.4 	& 96.4 	& 93.0 	& 96.4 	& 93.6 	& 96.3 	& 91.8 	& 95.4 	& 91.4  & 97.0 & 94.4\\
			ambulance        & 93.5 	& 90.1 	& \textbf{94.0} 	& \textbf{90.2} 	& 92.9 	& 89.0 	& 90.8 	& 85.4 	& 91.3 	& 86.5 	& 92.2 	& 88.1  & 92.1 & 87.9 \\
			ant & \textbf{92.1} & \textbf{91.6} 	& 88.9 	& 89.0 	& 90.3 	& 89.3 	& 91.0 	& 89.4 	& 85.9 	& 84.5 	& 87.1 	& 86.7   & 88.9 & 86.8 \\
			apple & 96.4 	& 90.7 	& \textbf{97.1} 	& \textbf{93.1} & 95.9 	& 89.9 	& 89.9 	& 73.7 	& 96.0 	& 88.7 	& 96.6 	& 90.4  & 95.8 & 88.8 \\ 
			... & ...& ...& ...&...&...&...&...&...&...&...&...&  ...& ...& \\\hline
			\rule{0pt}{10pt}
			\textbf{Average} & \textbf{96.7} 	& \textbf{94.5} 	& 96.3 	& 93.8 	& 95.7 	& 92.9 	& 95.3 	& 92.0 	& 95.5 	& 92.3 	& 95.5 	& 92.7 & 95.6 & 92.5\\ \hline
			\\
		\end{tabular}
		\label{tbl:gcns}
	\end{table*}

	\begin{table}[]
		\caption{Evaluation on different numbers of GCN units.}
		\centering
		\begin{tabular}{|c|c|c|c|c|}
			\hline
			\textbf{Average}   & \textbf{4 units} & \textbf{6 units} & \textbf{8 units} & \textbf{10 units} \\ \hline
			\textbf{P\_metric} & 97.6\%    & 96.9\%    & 96.9\%    & 96.7\%     \\ \hline
			\textbf{C\_metric} & 95.6\%    & 94.2\%    & 94.2\%    & 94.0\%     \\ \hline
		\end{tabular}
		\label{tbl:nblocks}
	\end{table}
	
	\begin{table}[]
		\caption{Evaluation on different values of N.}
		\centering
		\begin{tabular}{|c|c|c|c|}
			\hline
			\textbf{Average}   & \textbf{128} & \textbf{256}     & \textbf{512}     \\ \hline
			\textbf{P\_metric} & 95.97\%      & 96.62\%          & \textbf{96.69\%} \\ \hline
			\textbf{C\_metric} & 93.67\%      & \textbf{94.44\%} & 94.36\%          \\ \hline
		\end{tabular}
		\label{tbl:n}
	\end{table}	
	
	\subsection{Ablation Study}\label{subsec:ablation}
	In this section we examine the effectiveness of our various design choices in SketchGNN. All the experiments are run on the SPG dataset since it has sufficient data and complexity. We use 650 sketches per category to train the models and choose the optimal model with the lowest average loss on the validation set with 50 sketches in 100 epoch training. The remaining 100 sketches are  used as a test set to show the final performance.
	
	\paragraph{Structure Design}
	We use the network with only the DConv units and the max-pooling aggregation as the baseline, which is a typical single-branch two-level-feature structure, similar to the network used in \cite{Wang:2019:DGCNN}.
	Compared to the baseline, our model has two core improvements in structural design: 1) from two-level-feature to three-level-feature by adding the  stroke-pooling operation; 2) from one-branch to two-branch by using both the static and dynamic graph convolutions. 
	
	Table \ref{tbl:albation2} shows the benefit of these two designs:
	1) With the stroke-pooling (i.e., S.P.), \textbf{Base + S.P.} outperforms the baseline by 6.8\% in the pixel metric and 9.73\% in the component metric;
	2) With an additional static branch to generate the point-level features, \textbf{Base + S.D.} outperforms the baseline by 2.74\% in the pixel metric and 3.06\% in the component metric.
	To show the effectiveness of the two-branch structure we use, we design another two alternatives: the baseline with an additional dynamic branch (\textbf{Base+D.D.}) and the baseline with a static branch and a dynamic branch but use the SConv blocks before the pooling operation (\textbf{Base+D.S.}). Both of these networks are worse than \textbf{Base+S.D.}.
	Therefore, we use the baseline with an additional static branch and stroke-pooling as our full model (\textbf{Base + S.D. + S.P.}). Our full model achieves the best results: 7.48\% better than the baseline in the pixel metric and 10.86\% in the component metric.

	\paragraph{{GNN} variants}
	For the choice of the convolution operation, we compare the effects of various {GNN} variants, including EdgeConv \cite{Wang:2019:DGCNN}, MRGCN \cite{Li:2019:DGCN}, GraphSAGE \cite{Hamilton2017Inductive}, GIN \cite{xu2018powerful}, and ECC \cite{Martin:2017:DEDF}. 
	For ECC, we use $E(j, i) = ({\delta}_x, {\delta}_y,\|\delta\| , \arctan {\delta}_y/{\delta}_x ) $ as its input edge features, where $\delta = p_j - p_i$ is the offset between two nodes, similar to their original design.
	Table \ref{tbl:gcns} shows the experiments for the first 5 categories and the average results of the 20 categories on the SPG dataset. We use EdgeConv in our final model for its best performance in the experiments.	
	
	\begin{figure}
		\begin{center}		
			\includegraphics[width=\linewidth]{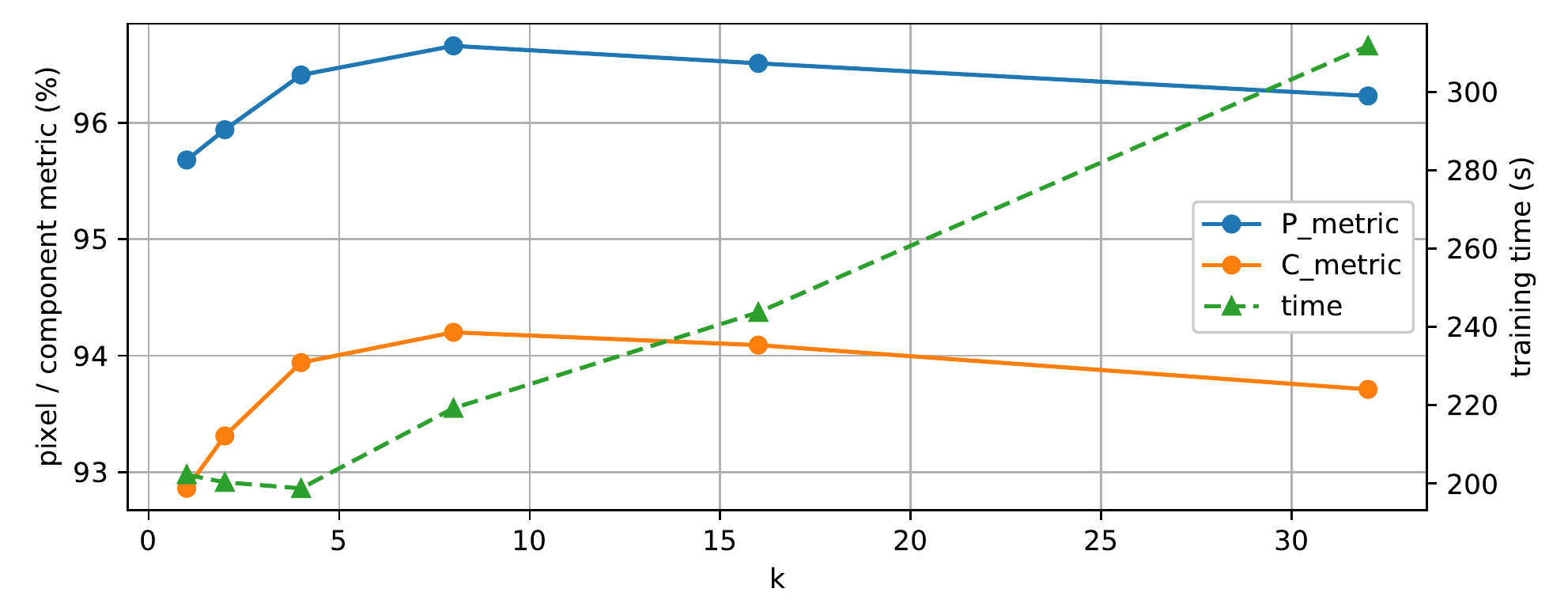}
		\end{center}
		\vspace{-3mm}
		\caption{Evaluation of accuracy and efficiency with different values of K. The solid lines represent the evaluation results of two accuracy metrics. The dashed line represents the training time cost with different values of $K$.}
		\label{fig:k}
	\end{figure}

	\paragraph{The Number of GCN Units}
	We have tried alternating the number of GCN units used in our two branches. In our current setting, we use both 4 units (of \emph{SConv} and \emph{DConv}) in the global and the local branches. Alternatively, we change this number to 6, 8, and 10 convolutional units. As shown in Table \ref{tbl:nblocks},  we find that increasing the number of GCN units does not benefit our results. {For simplicity,} we thus use 4 units in our final model.
	
	\paragraph{The Number of Sample Points}
	We choose $N=256$ based on the complexity of {sketches} in the SPG dataset. With $N=256$, we will not lose the details of the original sketches. To investigate the impact of $N$, we also train the model with $N=128$ and $N=512$. Table \ref{tbl:n} shows that {using} more points {does not} bring a significant improvement, while using fewer points reduces the performance.
	
	\paragraph{The Value of K}
	In consideration of accuracy
	and efficiency, we choose $K=8$ when we build dynamic edges by the dilated KNN function. Generally, the larger the value of $K$, the longer the running time of the KNN algorithm. Fig. \ref{fig:k} shows that $K=8$
	strikes a great balance between accuracy and efficiency. 
	Note that the segmentation accuracy starts to drop gradually when $K$ is larger 8,  possibly because the increased connection between the nodes causes an oversmoothing of the learned features.

	
	\paragraph{Classification Test}\label{subsec:classify}
	In an interesting attempt, we try to use our model for the sketch recognition task with minor modifications. After we aggregate the features for every node according to Eq. \ref{eqn:features}, we apply MLP and average pooling on every point of sketches to get graph-level features. Such graph-level features are used to classify sketches by an MLP classifier. We run our experiments on the TU-Berlin dataset with 250 classes. However, the effect is not satisfying, with only {a recognition accuracy} of 52.3\%. 
	We suspect that the poor performance may be due to the different scales of the segmentation and recognition problems. According to Table \ref{tbl:static}, the maximum number of segmentation labels is 11, while the number of classification categories in TU-Berlin is 250. The model we design specifically for semantic segmentation might not have sufficient capacity to extract enough discriminating feature information for large-scale classification. 
	{We also perform the same classification experiment using the DGCNN model \cite{Wang:2019:DGCNN} with the default settings (except for $k=8$ to keep it the same as in our test), and the resulting classification accuracy is only 47.3\%.}
	{The existing GNNs may have difficulty handling the sketch classification task for the sparse structure {of sketches}.}
	Hence, for such tasks, further exploration is needed with GNNs in future work.
	
\begin{figure}[t]
		\begin{center}		
			\includegraphics[width=\linewidth]{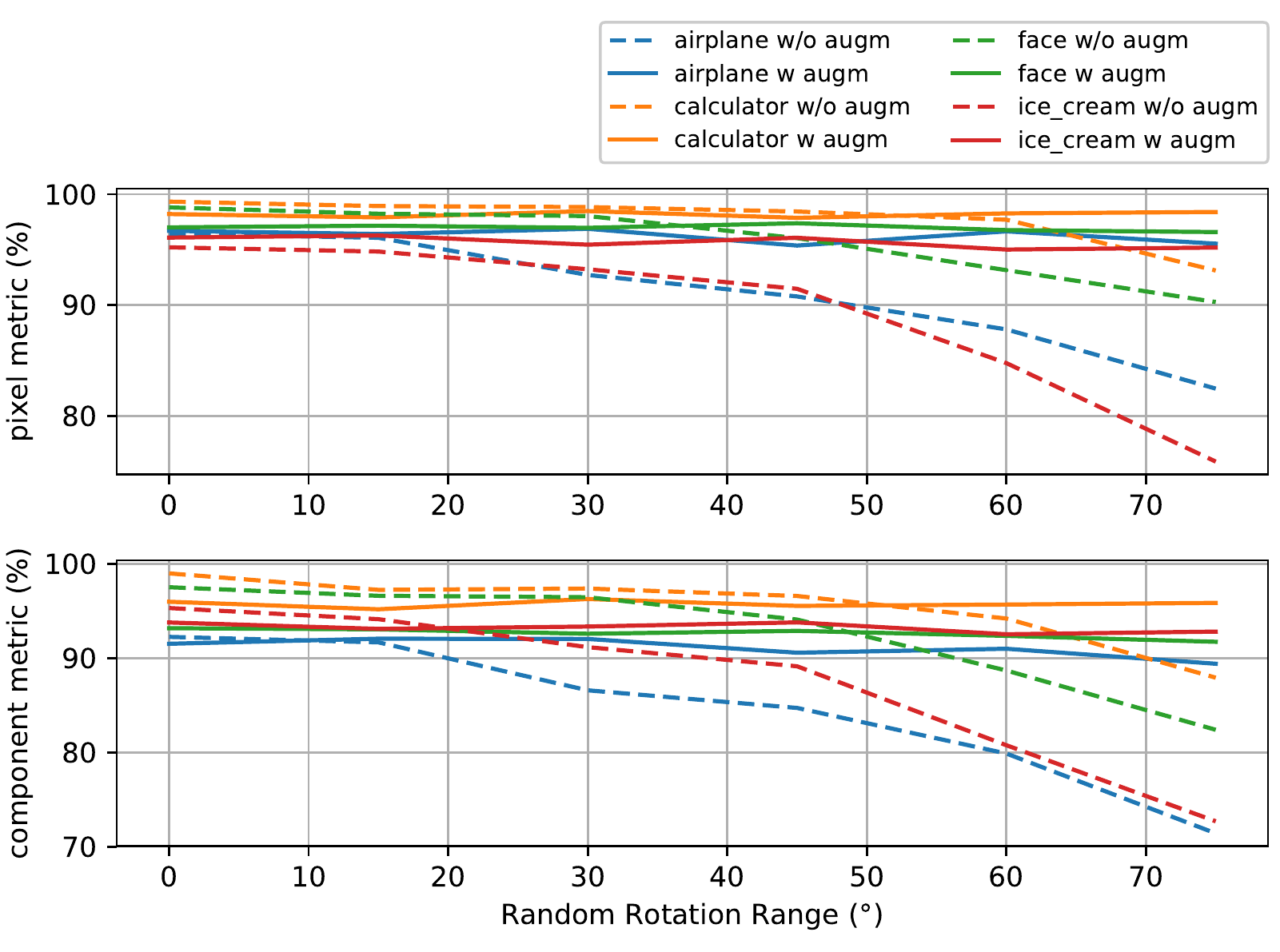}
		\end{center}
		\begin{center}		
			\includegraphics[width=\linewidth]{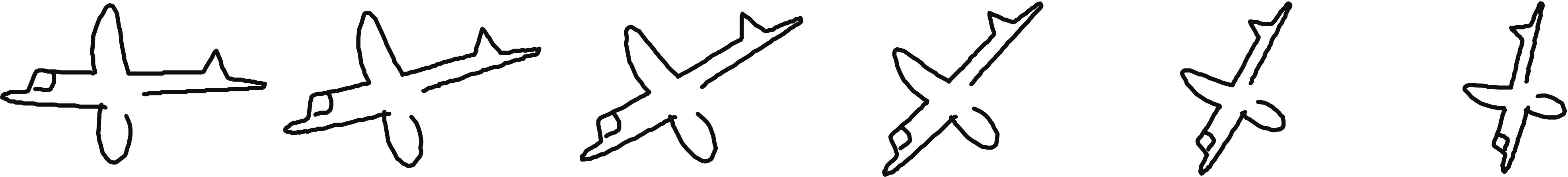}
		\end{center}
		\begin{center}		
			\includegraphics[width=\linewidth]{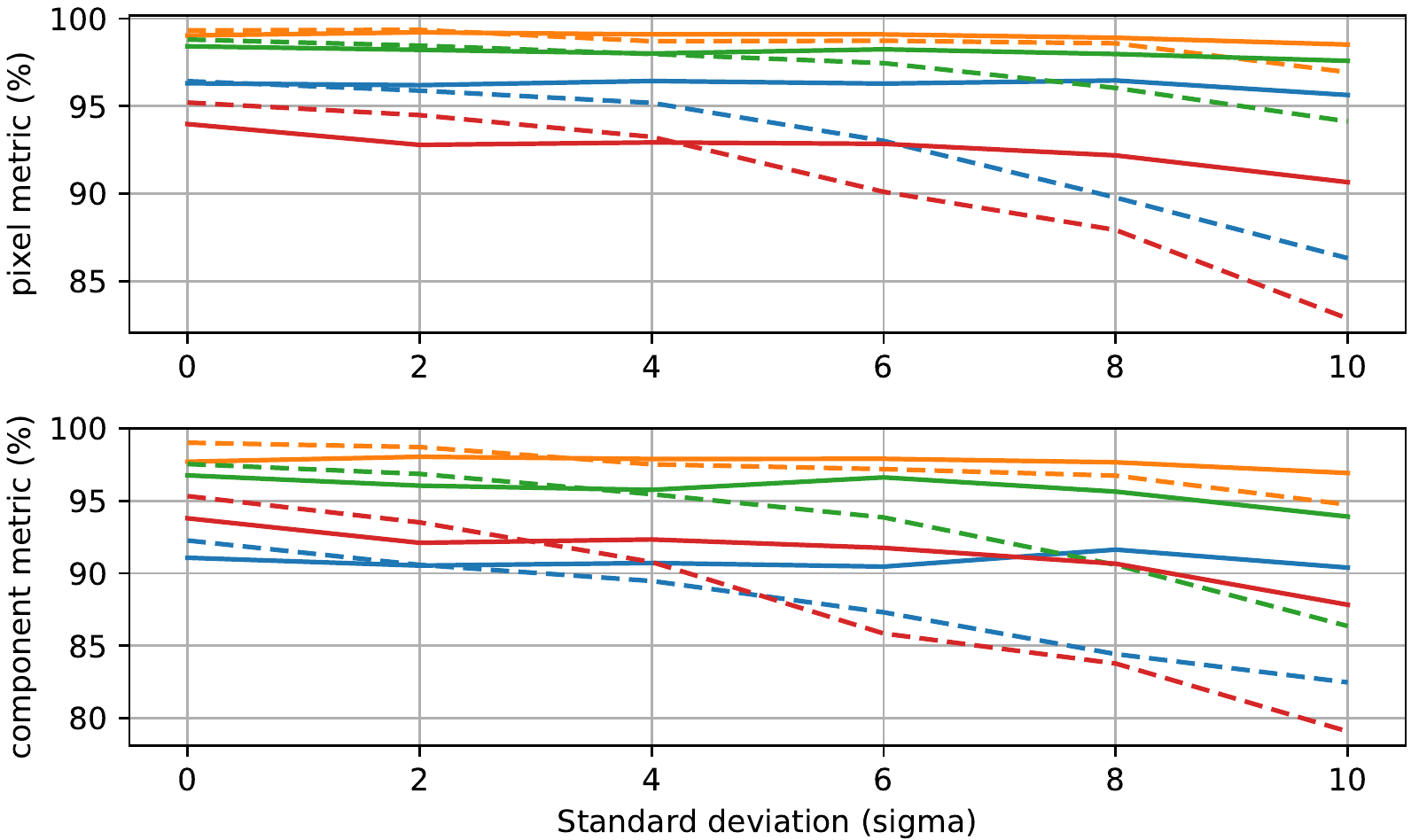}
		\end{center}
		\begin{center}		
			\includegraphics[width=\linewidth]{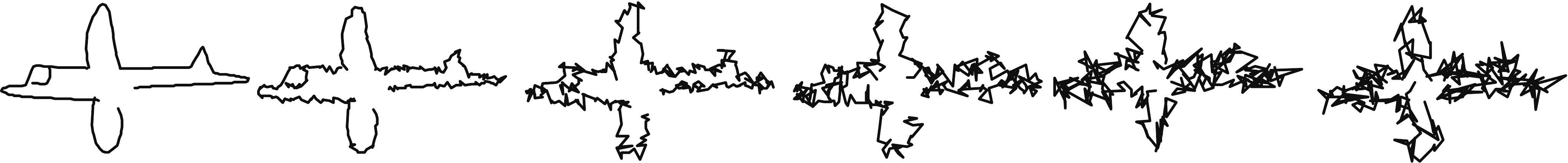}
		\end{center}
		\caption{The visualization of {the sketch-level invariance test (Top) and the point-level invariance test (Bottom)} on four representative categories. The solid lines represent the evaluation results with data augmentation in the training stage, while the dashed lines represents those without data augmentation.}
		\label{fig:permutation}
	\end{figure}

\begin{figure}[t]
		\begin{center}		
			\includegraphics[width=\linewidth]{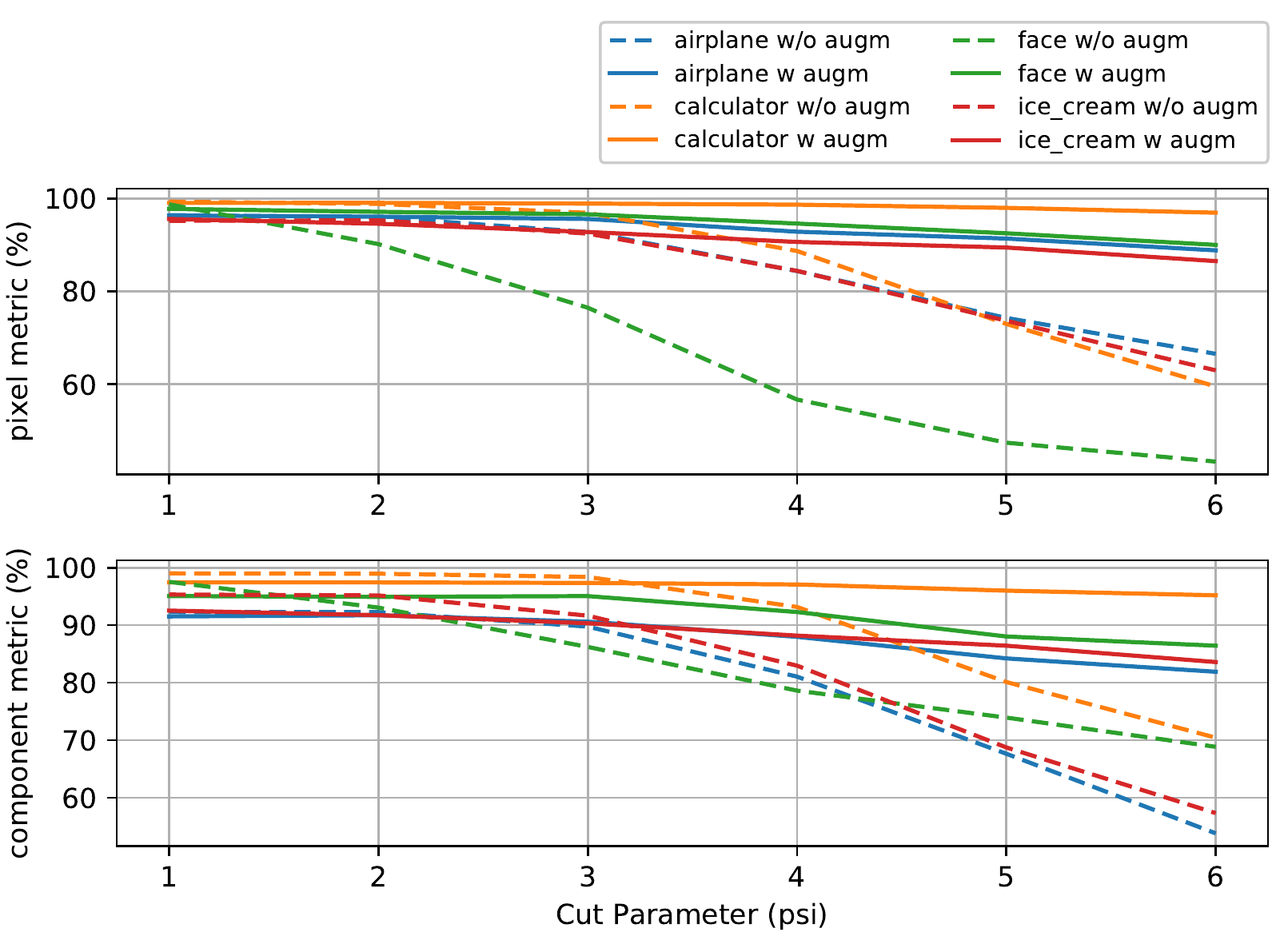}
		\end{center}
		\begin{center}		
			\includegraphics[width=\linewidth]{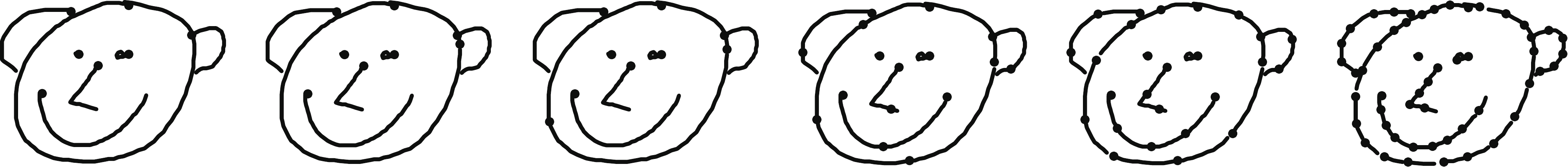}
		\end{center}
		\begin{center}		
			\includegraphics[width=\linewidth]{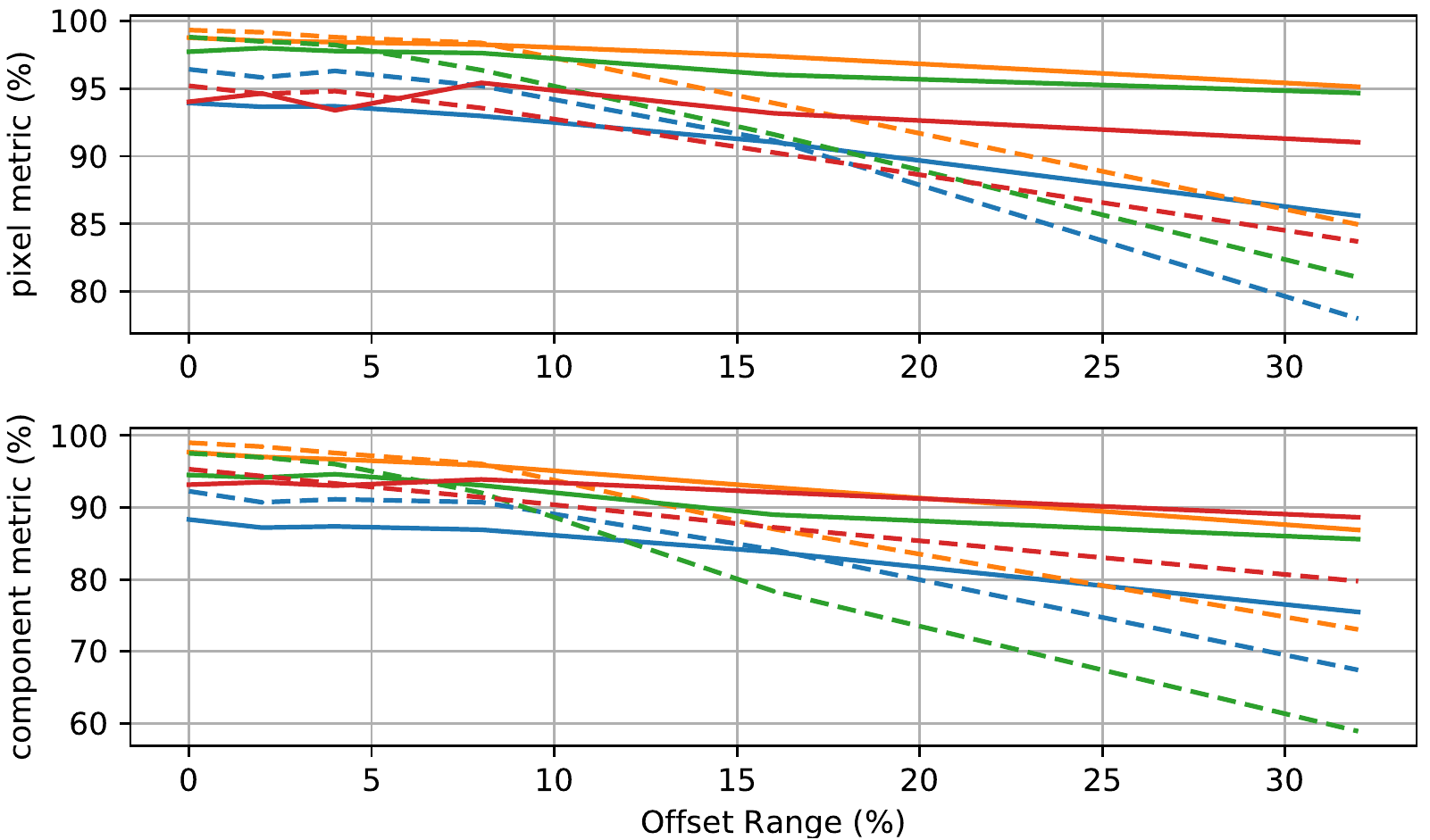}
		\end{center}
		\begin{center}		
			\includegraphics[width=\linewidth]{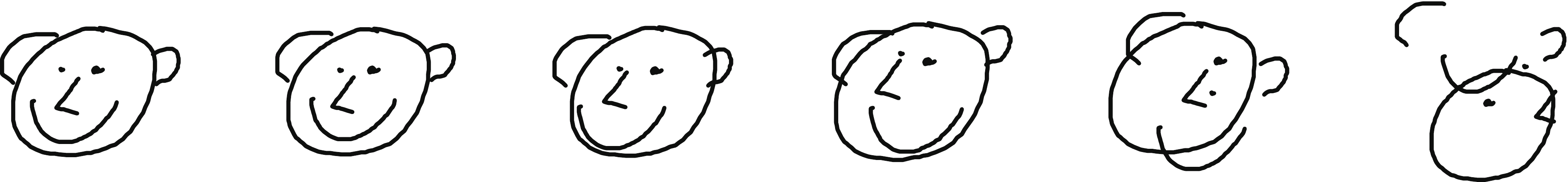}
		\end{center}
		\caption{{The visualization of two stroke-level invariance tests (Top for {test} \uppercase\expandafter{\romannumeral1} and Bottom for {test} \uppercase\expandafter{\romannumeral2}) on four representative categories. The solid lines represent the evaluation results with data augmentation in the training stage, while the dashed lines represents those without data augmentation.}}
		\label{fig:permutation2}
	\end{figure}

\begin{figure}[h]
		\begin{center}		
			\includegraphics[width=\linewidth]{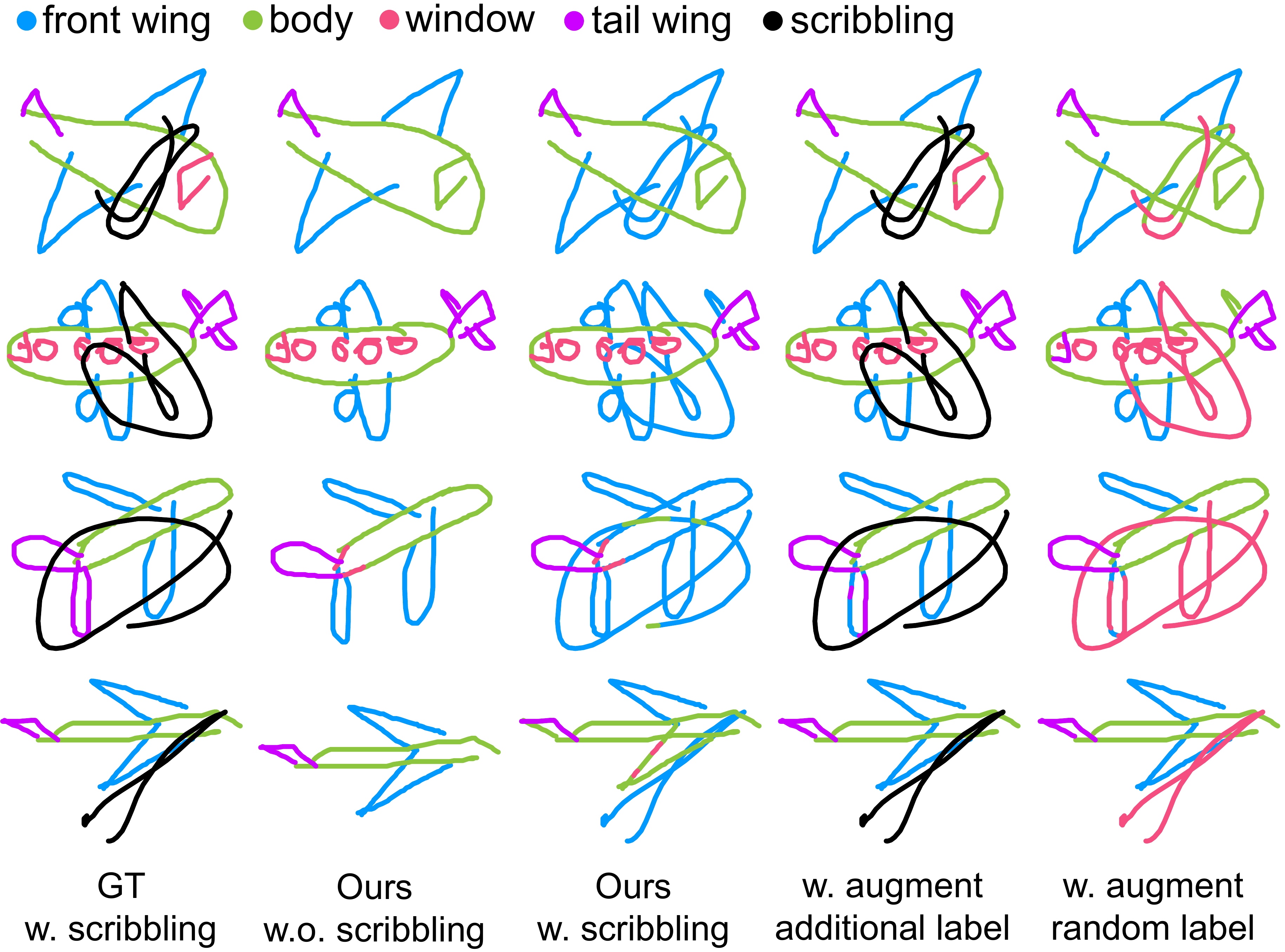}
		\end{center}
		\caption{{Representative results of stroke-level invariance test \uppercase\expandafter{\romannumeral3} on Category Airplane.}}
		\label{fig:perm_meanless}
	\end{figure}

	{
		
		\subsection{Invariance Test}\label{subsec:invariance}
		
		To further demonstrate our method, we design invariance tests {at} 
		three levels, i.e., sketch-level, stroke-level, and point-level. We perform the tests on four representative categories, i.e., Airplane, Calculator, Face, and Ice cream.
		
		\paragraph{Sketch-level} 
		Since we scale all {sketches} 
		to fit a $256\times256$ canvas, our method is not sensitive to translation or scaling. For {the} sketch-level invariance test, we apply random rotations to the sketches and monitor the segmentation results. Fig. \ref{fig:permutation} (Top) shows the results. It can be seen that as the range of the rotation angle increases, the performance decreases (dashed lines) rapidly if the training data does not contain similar sketches with random rotations. After we augment the training data with random rotations, the segmentation (solid lines) becomes more robust.
		
		\paragraph{Point-level} 
		For {the} point-level invariance test, we corrupt the input sketches with Gaussian noise. {Specifically,} we add random offset $O_p=(x_{o_p}, y_{o_p})$ to every point in the sketch, where $x_{o_p}, y_{o_p} \sim N(0, \delta^2),  \delta=0,2,4,6,8,10$. It can be seen from Fig. \ref{fig:permutation} (Bottom) that training our model on the model with similar random noise can greatly improve the robustness of our method to random noise.
		
		\paragraph{Stroke-level}
		For {the} stroke-level invariance test, we design three experiments. 
		
		In test \uppercase\expandafter{\romannumeral1}, we break the original strokes into small pieces. {This changes} 
		the relationship of the points (the input graph)
		{but not} 
		spatial layouts (i.e., the position) of the points (the input features). 
		After partition, each new stroke contains up to $p_s$ vertices, where $p_s=\frac{10N}{ 2^{\psi}\cdot n_s} $, $\psi = 1,2,3,4,5,6$, $N$ is the number of points in the sketch, {and} $n_s$ is the number of strokes in the sketch. Fig. \ref{fig:permutation2} (Top) shows the results. 
		The partition destroys the original stroke structure, causing poor segmentation results. When chopping up the strokes ($\psi = 6$), the results become disastrous. Relevant data augmentation can help our model adapt to the broken strokes, as shown in Fig. \ref{fig:permutation2}. However, it cannot make the model extremely robust to such topological changes since the destroy of the stroke structure would undermine the benefit of the static-graph branch and stroke-pooling.
		
		In test \uppercase\expandafter{\romannumeral2}, we add the random offset $O_s=(x_{o_s}, y_{o_s})$ to every stroke in the sketch, where $x_{o_s}, y_{o_s} \sim U(-\eta\times 256, \eta\times256)$, $\eta=2\%, 4\%, 8\%, 16\%, 32\%$. In this setting, the position of points (the input features) has changed, but the relationship of the points (the input graph) {remains the same}. 
		Fig.\ref{fig:permutation2} (Bottom) shows that data augmentation can improve the results but cannot eliminate the negative impact caused by the random offset since the random offset of the strokes can disrupt the learning of the relationship between the strokes. The experiment also shows that the correct inter-stroke relationships can benefit the segmentation task, and our model can learn such relationships.
		
		\begin{table}[b]
			\caption{{The component metric of our model and our model with two different data augmentation strategies on four representative categories.}}
			\begin{tabular}{c|ccc|ccc}
				\hline
				\multirow{2}{*}{\textbf{C\_metric}} & \multicolumn{3}{c|}{test w.o. scribbling}             & \multicolumn{3}{c}{test w. scribbling}               \\ 
				\textbf{}  & \textbf{Ours} & \textbf{Ours*} & \textbf{Ours**} & \textbf{Ours} & \textbf{Ours*} & \textbf{Ours**} \\\hline
				airplane   & 92.3          & 93.1           & \textbf{93.5}  & 82.6          & \textbf{93.9}  & 91.8           \\
				calculator & \textbf{99.0} & 97.4           & 98.3           & 96.0          & \textbf{97.7}  & 97.5           \\
				face       & \textbf{97.5} & 96.5           & 96.2           & 94.1          & \textbf{96.7}  & 94.8           \\
				ice cream & \textbf{95.3} & 91.1           & 92.2           & 90.5          & \textbf{93.3}  & 93.0 \\ \hline        
			\end{tabular}
			\label{tbl:scribble}
		\end{table}
		
		In test \uppercase\expandafter{\romannumeral3}, we scribble the sketch with meaningless strokes, leading to a decrease in the segmentation accuracy. Considering that re-sampling the sketch with scribbling stroke will change the original vertex position, we only use \textbf{C\_metric} for quantitative analysis here. Some categories are not severely affected by scribbling, e.g., the Calculator (96.0\% in C\_metric, -3.0\%). Some are severely affected, e.g., Airplane (82.6\% in C\_metric, -9.7\%). We experiment with two strategies of data augmentation to combat the scribbling. One is to set another new label for the meaningless strokes (\textbf{Ours*}), and the other is to randomly assign one of the existing labelings to the meaningless strokes (\textbf{Ours**}) during the training stage. Table \ref{tbl:scribble} shows that both strategies can work. On Category Airplane, the data augmentation even benefit{s} the segmentation. Fig. \ref{fig:perm_meanless} visualizes some segmentation results on Category Airplane in this experiment.
		
	}

	\begin{figure}[]
		\centering
		\includegraphics[width=\linewidth]{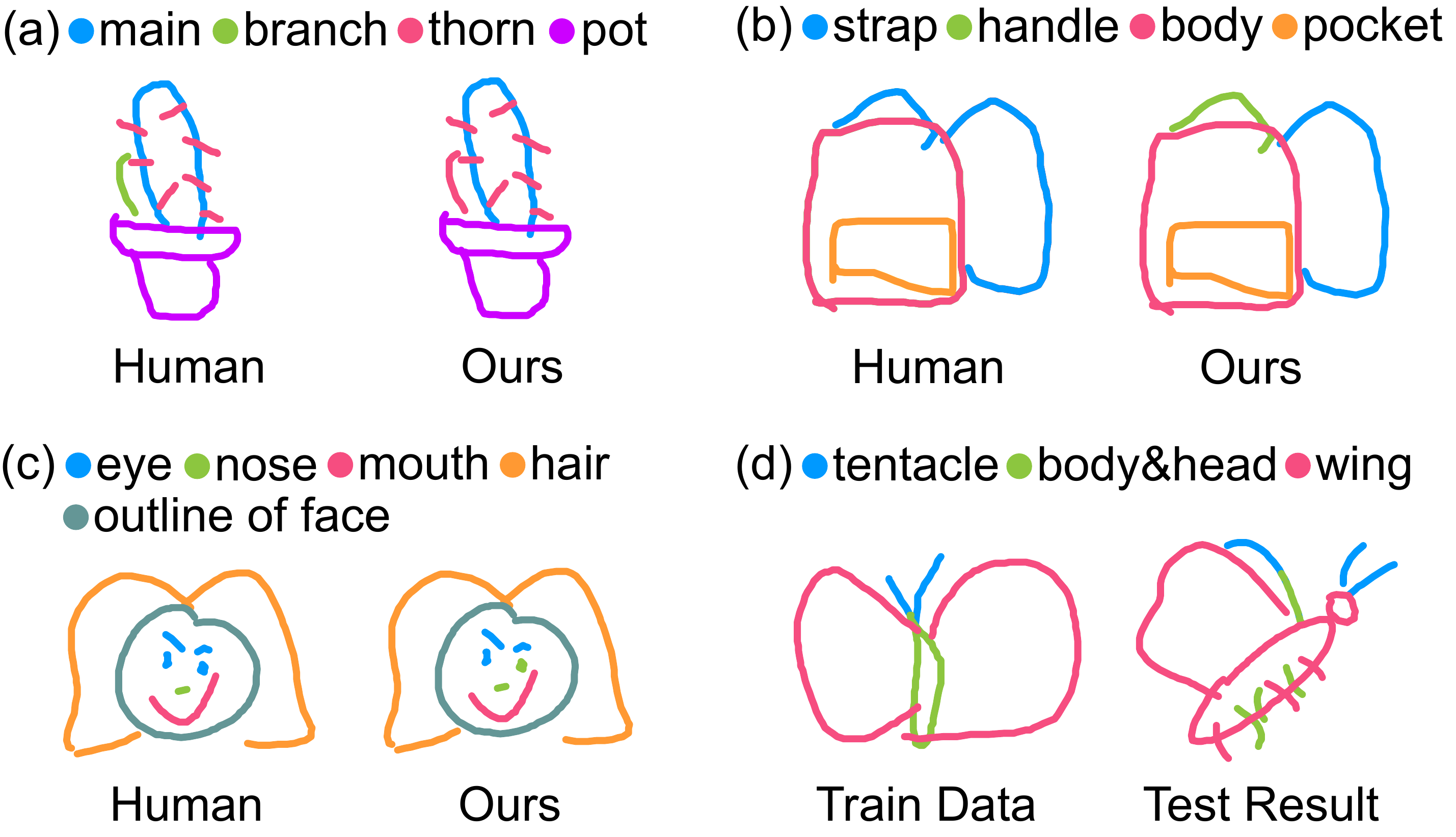}
		\caption{Exemplar results with imperfect segmentation.}
		\label{fig:failure cases}
	\end{figure}
	
	\section{Limitations}
	Fig.~\ref{fig:failure cases} shows several {segmentation results with segmentation errors.}
	The imperfection of our method is mainly caused by two factors. First, due to the inherent ambiguity of freehand sketches in part position and part shape, our model may assign wrong labels to strokes. For example, in Fig.~\ref{fig:failure cases} {(a) the branch of the cactus is mistakenly assigned as thorn and (b) the strap on the top {of a bag} is labeled as handle}. Second, the large differences between train data and test data may also mislead our model (Fig.~\ref{fig:failure cases} (d)): the butterflies in the train data always spread the wings, while the test example in this figure has the butterfly folded its wings, {with a different view angle}. We believe that the domain gap is a common issue of current learning based methods. Nevertheless, the visual results in Fig. \ref{fig:res} and the statistics on the Huang14 and the TU-Berlin datasets (Tables \ref{tbl:Huang14} and \ref{tbl:TUB}) have shown the generalization ability of our model. Finally, since our graph representation only warps features such as node position and proximity, our model is not aware of some high-level semantics such as the fact that ``a human face can only have one nose'' (see Fig.~\ref{fig:failure cases} (c)). We believe this issue can be alleviated by incorporating more semantic features into our graph representation for which we leave for future work.	
	
	\section{Conclusion}
	
	In this work, we presented the first graph convolutional network for semantic sketch segmentation and labeling. Our SketchGNN employs static graph convolutional units and dynamic graph convolutional units to {respectively} extract intra-stroke and inter-stroke features using a two-branch architecture. With a novel stroke pooling operation enabling more consistent intra-stroke labeling, our method achieves {higher accuracy than the state-of-the-art methods} with significantly fewer parameters in multiple sketch datasets. In our current experiments, we only use absolute positions as graph node features, while ignoring the information of the stroke order, direction, spatial relation, etc. In the future, we will exploit these information with more flexible graph structures. Another possibility would be to exploit recurrent modules to learn intact graph representations. Finally, it could be an intriguing direction to reshape our architecture for scene-level sketch segmentation and sketch recognition tasks.

	\begin{acks}
		We would like to thank the anonymous reviewers for their constructive comments. This work is supported in part by the National Key Research \& Development Program of China (2018YFE0100900). Hongbo Fu was supported by unrestricted gifts from Adobe and grants from the Research Grants Council of the Hong Kong Special Administrative Region, China (No. CityU 11212119, CityU 11237116), City University of Hong Kong (No. 7005176), and the Centre for Applied Computing and Interactive Media (ACIM) of School of Creative Media, CityU. 
	\end{acks}
	
	\bibliographystyle{ACM-Reference-Format}
	\bibliography{sample-base}


\begin{thebibliography}{47}


\ifx \showCODEN    \undefined \def \showCODEN     #1{\unskip}     \fi
\ifx \showDOI      \undefined \def \showDOI       #1{#1}\fi
\ifx \showISBNx    \undefined \def \showISBNx     #1{\unskip}     \fi
\ifx \showISBNxiii \undefined \def \showISBNxiii  #1{\unskip}     \fi
\ifx \showISSN     \undefined \def \showISSN      #1{\unskip}     \fi
\ifx \showLCCN     \undefined \def \showLCCN      #1{\unskip}     \fi
\ifx \shownote     \undefined \def \shownote      #1{#1}          \fi
\ifx \showarticletitle \undefined \def \showarticletitle #1{#1}   \fi
\ifx \showURL      \undefined \def \showURL       {\relax}        \fi
\providecommand\bibfield[2]{#2}
\providecommand\bibinfo[2]{#2}
\providecommand\natexlab[1]{#1}
\providecommand\showeprint[2][]{arXiv:#2}

\bibitem[\protect\citeauthoryear{Bastings, Titov, Aziz, Marcheggiani, and
  Sima{'}an}{Bastings et~al\mbox{.}}{2017}]%
        {Bastings:2017:GCESNMT}
\bibfield{author}{\bibinfo{person}{Joost Bastings}, \bibinfo{person}{Ivan
  Titov}, \bibinfo{person}{Wilker Aziz}, \bibinfo{person}{Diego Marcheggiani},
  {and} \bibinfo{person}{Khalil Sima{'}an}.} \bibinfo{year}{2017}\natexlab{}.
\newblock \showarticletitle{Graph Convolutional Encoders for Syntax-aware
  Neural Machine Translation}. In \bibinfo{booktitle}{\emph{Proceedings of the
  2017 Conference on Empirical Methods in Natural Language Processing}}.
  \bibinfo{pages}{1957--1967}.
\newblock


\bibitem[\protect\citeauthoryear{Canny and John}{Canny and John}{1986}]%
        {Canny1986A}
\bibfield{author}{\bibinfo{person}{Canny} {and} \bibinfo{person}{John}.}
  \bibinfo{year}{1986}\natexlab{}.
\newblock \showarticletitle{A Computational Approach to Edge Detection}.
\newblock \bibinfo{journal}{\emph{Pattern Analysis and Machine Intelligence,
  IEEE Transactions on}} (\bibinfo{year}{1986}).
\newblock


\bibitem[\protect\citeauthoryear{Charles, Hao, Mo, and Guibas}{Charles
  et~al\mbox{.}}{2017}]%
        {Charles2017PointNet}
\bibfield{author}{\bibinfo{person}{R.~Qi Charles}, \bibinfo{person}{Su Hao},
  \bibinfo{person}{Kaichun Mo}, {and} \bibinfo{person}{Leonidas~J. Guibas}.}
  \bibinfo{year}{2017}\natexlab{}.
\newblock \showarticletitle{PointNet: Deep Learning on Point Sets for 3D
  Classification and Segmentation}. In \bibinfo{booktitle}{\emph{2017 IEEE
  Conference on Computer Vision and Pattern Recognition (CVPR)}}.
  \bibinfo{pages}{77--85}.
\newblock


\bibitem[\protect\citeauthoryear{Chen, Papandreou, Kokkinos, Murphy, and
  Yuille}{Chen et~al\mbox{.}}{2018}]%
        {deeplab2018}
\bibfield{author}{\bibinfo{person}{Liang{-}Chieh Chen}, \bibinfo{person}{George
  Papandreou}, \bibinfo{person}{Iasonas Kokkinos}, \bibinfo{person}{Kevin
  Murphy}, {and} \bibinfo{person}{Alan~L. Yuille}.}
  \bibinfo{year}{2018}\natexlab{}.
\newblock \showarticletitle{DeepLab: Semantic Image Segmentation with Deep
  Convolutional Nets, Atrous Convolution, and Fully Connected CRFs}.
\newblock \bibinfo{journal}{\emph{{IEEE} Trans. Pattern Anal. Mach. Intell.}}
  \bibinfo{volume}{40}, \bibinfo{number}{4} (\bibinfo{year}{2018}),
  \bibinfo{pages}{834--848}.
\newblock


\bibitem[\protect\citeauthoryear{Defferrard, Bresson, and
  Vandergheynst}{Defferrard et~al\mbox{.}}{2016}]%
        {2016Convolutional}
\bibfield{author}{\bibinfo{person}{Michal Defferrard}, \bibinfo{person}{Xavier
  Bresson}, {and} \bibinfo{person}{Pierre Vandergheynst}.}
  \bibinfo{year}{2016}\natexlab{}.
\newblock \showarticletitle{Convolutional Neural Networks on Graphs with Fast
  Localized Spectral Filtering}.
\newblock  (\bibinfo{year}{2016}).
\newblock


\bibitem[\protect\citeauthoryear{Delaye and Lee}{Delaye and Lee}{2015}]%
        {Delaye2015A}
\bibfield{author}{\bibinfo{person}{Adrien Delaye} {and} \bibinfo{person}{Kibok
  Lee}.} \bibinfo{year}{2015}\natexlab{}.
\newblock \showarticletitle{A flexible framework for online document
  segmentation by pairwise stroke distance learning.}
\newblock \bibinfo{journal}{\emph{Pattern Recognition}} \bibinfo{volume}{48},
  \bibinfo{number}{4} (\bibinfo{year}{2015}), \bibinfo{pages}{1197--1210}.
\newblock


\bibitem[\protect\citeauthoryear{Douglas and Peucker}{Douglas and
  Peucker}{1973}]%
        {Douglas1973Algorithms}
\bibfield{author}{\bibinfo{person}{David~H. Douglas} {and}
  \bibinfo{person}{Thomas~K. Peucker}.} \bibinfo{year}{1973}\natexlab{}.
\newblock \bibinfo{booktitle}{\emph{Algorithms for the reduction of the number
  of points required to represent a line or its caricature}}.
\newblock \bibinfo{publisher}{University of Toronto Press}.
\newblock


\bibitem[\protect\citeauthoryear{Eitz, Hays, and Alexa}{Eitz
  et~al\mbox{.}}{2012a}]%
        {eitz2012hdhso}
\bibfield{author}{\bibinfo{person}{Mathias Eitz}, \bibinfo{person}{James Hays},
  {and} \bibinfo{person}{Marc Alexa}.} \bibinfo{year}{2012}\natexlab{a}.
\newblock \showarticletitle{How Do Humans Sketch Objects?}
\newblock \bibinfo{journal}{\emph{ACM Transactions on Graphics (Proceedings
  SIGGRAPH)}} \bibinfo{volume}{31}, \bibinfo{number}{4} (\bibinfo{year}{2012}),
  \bibinfo{pages}{44:1--44:10}.
\newblock


\bibitem[\protect\citeauthoryear{Eitz, Richter, Boubekeur, Hildebrand, and
  Alexa}{Eitz et~al\mbox{.}}{2012b}]%
        {eitz2012sbsr}
\bibfield{author}{\bibinfo{person}{Mathias Eitz}, \bibinfo{person}{Ronald
  Richter}, \bibinfo{person}{Tamy Boubekeur}, \bibinfo{person}{Kristian
  Hildebrand}, {and} \bibinfo{person}{Marc Alexa}.}
  \bibinfo{year}{2012}\natexlab{b}.
\newblock \showarticletitle{Sketch-Based Shape Retrieval}.
\newblock \bibinfo{journal}{\emph{ACM Transactions on Graphics (Proceedings
  SIGGRAPH)}} \bibinfo{volume}{31}, \bibinfo{number}{4} (\bibinfo{year}{2012}),
  \bibinfo{pages}{31:1--31:10}.
\newblock


\bibitem[\protect\citeauthoryear{Gennari, Kara, Stahovich, and Shimada}{Gennari
  et~al\mbox{.}}{2005}]%
        {Gennari2005Combining}
\bibfield{author}{\bibinfo{person}{Leslie Gennari},
  \bibinfo{person}{Levent~Burak Kara}, \bibinfo{person}{Thomas~F. Stahovich},
  {and} \bibinfo{person}{Kenji Shimada}.} \bibinfo{year}{2005}\natexlab{}.
\newblock \showarticletitle{Combining geometry and domain knowledge to
  interpret hand-drawn diagrams}.
\newblock \bibinfo{journal}{\emph{Computers \& Graphics}} \bibinfo{volume}{29},
  \bibinfo{number}{4} (\bibinfo{year}{2005}), \bibinfo{pages}{547--562}.
\newblock


\bibitem[\protect\citeauthoryear{Ha and Eck}{Ha and Eck}{2017}]%
        {Ha2017A}
\bibfield{author}{\bibinfo{person}{David Ha} {and} \bibinfo{person}{Douglas
  Eck}.} \bibinfo{year}{2017}\natexlab{}.
\newblock \showarticletitle{A Neural Representation of Sketch Drawings}. In
  \bibinfo{booktitle}{\emph{6th International Conference on Learning
  Representations, {ICLR} 2018}}.
\newblock


\bibitem[\protect\citeauthoryear{Hamilton, Ying, and Leskovec}{Hamilton
  et~al\mbox{.}}{2017}]%
        {Hamilton2017Inductive}
\bibfield{author}{\bibinfo{person}{William~L. Hamilton}, \bibinfo{person}{Rex
  Ying}, {and} \bibinfo{person}{Jure Leskovec}.}
  \bibinfo{year}{2017}\natexlab{}.
\newblock \showarticletitle{Inductive Representation Learning on Large Graphs}.
  In \bibinfo{booktitle}{\emph{Proceedings of the 31st International Conference
  on Neural Information Processing Systems}}. \bibinfo{pages}{1025--1035}.
\newblock


\bibitem[\protect\citeauthoryear{Huang, Fu, and Lau}{Huang
  et~al\mbox{.}}{2014}]%
        {Huang:2014:DSL}
\bibfield{author}{\bibinfo{person}{Zhe Huang}, \bibinfo{person}{Hongbo Fu},
  {and} \bibinfo{person}{Rynson W.~H. Lau}.} \bibinfo{year}{2014}\natexlab{}.
\newblock \showarticletitle{Data-driven Segmentation and Labeling of Freehand
  Sketches}.
\newblock \bibinfo{journal}{\emph{ACM Trans. Graph.}} \bibinfo{volume}{33},
  \bibinfo{number}{6} (\bibinfo{year}{2014}), \bibinfo{pages}{175:1--175:10}.
\newblock


\bibitem[\protect\citeauthoryear{Kipf and Welling}{Kipf and Welling}{2016}]%
        {Thomas:2017:SCGCN}
\bibfield{author}{\bibinfo{person}{Thomas~N. Kipf} {and} \bibinfo{person}{Max
  Welling}.} \bibinfo{year}{2016}\natexlab{}.
\newblock \showarticletitle{Semi-Supervised Classification with Graph
  Convolutional Networks}. In \bibinfo{booktitle}{\emph{International
  Conference on Learning Representations}}.
\newblock


\bibitem[\protect\citeauthoryear{Li, Müller, Thabet, and Ghanem}{Li
  et~al\mbox{.}}{2019b}]%
        {Li:2019:DGCN}
\bibfield{author}{\bibinfo{person}{Guohao Li}, \bibinfo{person}{Matthias
  Müller}, \bibinfo{person}{Ali Thabet}, {and} \bibinfo{person}{Bernard
  Ghanem}.} \bibinfo{year}{2019}\natexlab{b}.
\newblock \showarticletitle{DeepGCNs: Can GCNs Go as Deep as CNNs?}. In
  \bibinfo{booktitle}{\emph{The IEEE International Conference on Computer
  Vision (ICCV)}}.
\newblock


\bibitem[\protect\citeauthoryear{Li, Pang, Song, Song, Xiang, Hospedales, and
  Zhang}{Li et~al\mbox{.}}{2018b}]%
        {Li:2018:USPG}
\bibfield{author}{\bibinfo{person}{Ke Li}, \bibinfo{person}{Kaiyue Pang},
  \bibinfo{person}{Jifei Song}, \bibinfo{person}{Yi-Zhe Song},
  \bibinfo{person}{Tao Xiang}, \bibinfo{person}{Timothy~M. Hospedales}, {and}
  \bibinfo{person}{Honggang Zhang}.} \bibinfo{year}{2018}\natexlab{b}.
\newblock \showarticletitle{Universal Sketch Perceptual Grouping}. In
  \bibinfo{booktitle}{\emph{The European Conference on Computer Vision
  (ECCV)}}. \bibinfo{pages}{582--597}.
\newblock


\bibitem[\protect\citeauthoryear{Li, Pang, Song, Xiang, Hospedales, and
  Zhang}{Li et~al\mbox{.}}{2019c}]%
        {Li:2019:TDUSPG}
\bibfield{author}{\bibinfo{person}{Ke Li}, \bibinfo{person}{Kaiyue Pang},
  \bibinfo{person}{Yi-Zhe Song}, \bibinfo{person}{Tao Xiang},
  \bibinfo{person}{Timothy~M. Hospedales}, {and} \bibinfo{person}{Honggang
  Zhang}.} \bibinfo{year}{2019}\natexlab{c}.
\newblock \showarticletitle{Toward Deep Universal Sketch Perceptual Grouper}.
\newblock \bibinfo{journal}{\emph{IEEE Transactions on Image Processing}}
  \bibinfo{volume}{28}, \bibinfo{number}{7} (\bibinfo{year}{2019}),
  \bibinfo{pages}{3219--3231}.
\newblock


\bibitem[\protect\citeauthoryear{Li, Fu, and Tai}{Li et~al\mbox{.}}{2019a}]%
        {Li:2019:Fast}
\bibfield{author}{\bibinfo{person}{Lei Li}, \bibinfo{person}{Hongbo Fu}, {and}
  \bibinfo{person}{Chiew-Lan Tai}.} \bibinfo{year}{2019}\natexlab{a}.
\newblock \showarticletitle{Fast Sketch Segmentation and Labeling With Deep
  Learning}.
\newblock \bibinfo{journal}{\emph{IEEE Computer Graphics and Applications}}
  \bibinfo{volume}{39}, \bibinfo{number}{2} (\bibinfo{year}{2019}),
  \bibinfo{pages}{38--51}.
\newblock


\bibitem[\protect\citeauthoryear{Li, Huang, Zou, Tai, Lau, Zhang, Tan, and
  Fu}{Li et~al\mbox{.}}{2016}]%
        {Lei2016Model}
\bibfield{author}{\bibinfo{person}{Lei Li}, \bibinfo{person}{Zhe Huang},
  \bibinfo{person}{Changqing Zou}, \bibinfo{person}{Chiew-Lan Tai},
  \bibinfo{person}{Rynson W.~H. Lau}, \bibinfo{person}{Hao Zhang},
  \bibinfo{person}{Ping Tan}, {and} \bibinfo{person}{Hongbo Fu}.}
  \bibinfo{year}{2016}\natexlab{}.
\newblock \showarticletitle{Model-driven sketch reconstruction with
  structure-oriented retrieval}. In \bibinfo{booktitle}{\emph{SIGGRAPH ASIA
  2016 Technical Briefs}}. \bibinfo{pages}{28:1--28:4}.
\newblock


\bibitem[\protect\citeauthoryear{Li, Zou, Zheng, Su, and Tai}{Li
  et~al\mbox{.}}{2020}]%
        {2020SketchR2CNN}
\bibfield{author}{\bibinfo{person}{Lei Li}, \bibinfo{person}{Changqing Zou},
  \bibinfo{person}{Youyi Zheng}, \bibinfo{person}{Qingkun Su}, {and}
  \bibinfo{person}{Chiew~Lan Tai}.} \bibinfo{year}{2020}\natexlab{}.
\newblock \showarticletitle{Sketch-R2CNN: An RNN-Rasterization-CNN Architecture
  for Vector Sketch Recognition}.
\newblock \bibinfo{journal}{\emph{IEEE Transactions on Visualization and
  Computer Graphics}} \bibinfo{volume}{PP}, \bibinfo{number}{99}
  (\bibinfo{year}{2020}), \bibinfo{pages}{1--1}.
\newblock


\bibitem[\protect\citeauthoryear{Li, Han, and Wu}{Li et~al\mbox{.}}{2018a}]%
        {Li:2018:DIGCNSL}
\bibfield{author}{\bibinfo{person}{Qimai Li}, \bibinfo{person}{Zhichao Han},
  {and} \bibinfo{person}{Xiao-Ming Wu}.} \bibinfo{year}{2018}\natexlab{a}.
\newblock \showarticletitle{Deeper Insights Into Graph Convolutional Networks
  for Semi-Supervised Learning}. In \bibinfo{booktitle}{\emph{Proceedings of
  the Thirty-Second {AAAI} Conference on Artificial Intelligence}}.
  \bibinfo{pages}{3538--3545}.
\newblock


\bibitem[\protect\citeauthoryear{Monti, Bronstein, and Bresson}{Monti
  et~al\mbox{.}}{2017}]%
        {Monti:2017:GMCRMNN}
\bibfield{author}{\bibinfo{person}{Federico Monti}, \bibinfo{person}{Michael
  Bronstein}, {and} \bibinfo{person}{Xavier Bresson}.}
  \bibinfo{year}{2017}\natexlab{}.
\newblock \showarticletitle{Geometric Matrix Completion with Recurrent
  Multi-Graph Neural Networks}. In \bibinfo{booktitle}{\emph{Advances in Neural
  Information Processing Systems 30}}. \bibinfo{pages}{3697--3707}.
\newblock


\bibitem[\protect\citeauthoryear{Noris, Sykora, Shamir, Coros, Whited, Simmons,
  Hornung, Gross, and Sumner}{Noris et~al\mbox{.}}{2012}]%
        {Noris2012SmartScribbles}
\bibfield{author}{\bibinfo{person}{Gioacchino Noris}, \bibinfo{person}{Daniel
  Sykora}, \bibinfo{person}{A. Shamir}, \bibinfo{person}{Stelian Coros},
  \bibinfo{person}{Brian Whited}, \bibinfo{person}{Maryann Simmons},
  \bibinfo{person}{Alexander Hornung}, \bibinfo{person}{Markus~H. Gross}, {and}
  \bibinfo{person}{Robert~W. Sumner}.} \bibinfo{year}{2012}\natexlab{}.
\newblock \showarticletitle{Smart Scribbles for Sketch Segmentation}.
\newblock \bibinfo{journal}{\emph{Computer Graphics Forum}}
  \bibinfo{volume}{31}, \bibinfo{number}{8} (\bibinfo{date}{January}
  \bibinfo{year}{2012}).
\newblock


\bibitem[\protect\citeauthoryear{Perteneder, Bresler, Grossauer, Leong, and
  Haller}{Perteneder et~al\mbox{.}}{2015}]%
        {Perteneder2015cLuster}
\bibfield{author}{\bibinfo{person}{Florian Perteneder}, \bibinfo{person}{Martin
  Bresler}, \bibinfo{person}{Eva~Maria Grossauer}, \bibinfo{person}{Joanne
  Leong}, {and} \bibinfo{person}{Michael Haller}.}
  \bibinfo{year}{2015}\natexlab{}.
\newblock \showarticletitle{cLuster: Smart Clustering of Free-Hand Sketches on
  Large Interactive Surfaces}. In \bibinfo{booktitle}{\emph{Proceedings of the
  28th Annual ACM Symposium on User Interface Software and Technology}}.
  \bibinfo{pages}{37--46}.
\newblock


\bibitem[\protect\citeauthoryear{{Qi}, {Guo}, {Li}, {Zhang}, {Xiang}, and
  {Song}}{{Qi} et~al\mbox{.}}{2013}]%
        {Qi:2013:SPG}
\bibfield{author}{\bibinfo{person}{Yonggang {Qi}}, \bibinfo{person}{Jun {Guo}},
  \bibinfo{person}{Yi {Li}}, \bibinfo{person}{Honggang {Zhang}},
  \bibinfo{person}{Tao {Xiang}}, {and} \bibinfo{person}{Yi-Zhe {Song}}.}
  \bibinfo{year}{2013}\natexlab{}.
\newblock \showarticletitle{Sketching by perceptual grouping}. In
  \bibinfo{booktitle}{\emph{2013 IEEE International Conference on Image
  Processing}}. \bibinfo{pages}{270--274}.
\newblock


\bibitem[\protect\citeauthoryear{Qi, Guo, Song, Xiang, Zhang, and Tan}{Qi
  et~al\mbox{.}}{2015}]%
        {Qi:2015:Im2Sketch}
\bibfield{author}{\bibinfo{person}{Yonggang Qi}, \bibinfo{person}{Jun Guo},
  \bibinfo{person}{Yi-Zhe Song}, \bibinfo{person}{Tao Xiang},
  \bibinfo{person}{Honggang Zhang}, {and} \bibinfo{person}{Zheng-Hua Tan}.}
  \bibinfo{year}{2015}\natexlab{}.
\newblock \showarticletitle{Im2Sketch: Sketch generation by unconflicted
  perceptual grouping}.
\newblock \bibinfo{journal}{\emph{Neurocomputing}}  \bibinfo{volume}{165}
  (\bibinfo{year}{2015}), \bibinfo{pages}{338--349}.
\newblock


\bibitem[\protect\citeauthoryear{{Qi}, {Song}, {Xiang}, {Zhang}, {Hospedales},
  {Li}, and {Guo}}{{Qi} et~al\mbox{.}}{2015}]%
        {Qi:2015:MBUOEVPG}
\bibfield{author}{\bibinfo{person}{Yonggang {Qi}}, \bibinfo{person}{Yi-Zhe
  {Song}}, \bibinfo{person}{Tao {Xiang}}, \bibinfo{person}{Honggang {Zhang}},
  \bibinfo{person}{Timothy {Hospedales}}, \bibinfo{person}{Yi {Li}}, {and}
  \bibinfo{person}{Jun {Guo}}.} \bibinfo{year}{2015}\natexlab{}.
\newblock \showarticletitle{Making Better Use of Edges via Perceptual
  Grouping}. In \bibinfo{booktitle}{\emph{2015 IEEE Conference on Computer
  Vision and Pattern Recognition (CVPR)}}. \bibinfo{pages}{1856--1865}.
\newblock


\bibitem[\protect\citeauthoryear{{Qi} and {Tan}}{{Qi} and {Tan}}{2019}]%
        {Qi:2019:SSegP}
\bibfield{author}{\bibinfo{person}{Yonggang {Qi}} {and}
  \bibinfo{person}{Zheng-Hua {Tan}}.} \bibinfo{year}{2019}\natexlab{}.
\newblock \showarticletitle{SketchSegNet+: An End-to-End Learning of RNN for
  Multi-Class Sketch Semantic Segmentation}. In \bibinfo{booktitle}{\emph{IEEE
  Access}}, Vol.~\bibinfo{volume}{7}. \bibinfo{pages}{102717--102726}.
\newblock


\bibitem[\protect\citeauthoryear{Sangkloy, Burnell, Ham, and Hays}{Sangkloy
  et~al\mbox{.}}{2016}]%
        {Sangkloy:2016:SDL}
\bibfield{author}{\bibinfo{person}{Patsorn Sangkloy}, \bibinfo{person}{Nathan
  Burnell}, \bibinfo{person}{Cusuh Ham}, {and} \bibinfo{person}{James Hays}.}
  \bibinfo{year}{2016}\natexlab{}.
\newblock \showarticletitle{The Sketchy Database: Learning to Retrieve Badly
  Drawn Bunnies}.
\newblock \bibinfo{journal}{\emph{ACM Trans. Graph.}} \bibinfo{volume}{35},
  \bibinfo{number}{4} (\bibinfo{year}{2016}), \bibinfo{pages}{119:1--119:12}.
\newblock


\bibitem[\protect\citeauthoryear{Sarvadevabhatla, Dwivedi, Biswas, Manocha, and
  Babu}{Sarvadevabhatla et~al\mbox{.}}{2017}]%
        {Sarvadevabhatla:2017:SketchParse}
\bibfield{author}{\bibinfo{person}{Ravi~Kiran Sarvadevabhatla},
  \bibinfo{person}{Isht Dwivedi}, \bibinfo{person}{Abhijat Biswas},
  \bibinfo{person}{Sahil Manocha}, {and} \bibinfo{person}{R.~Venkatesh Babu}.}
  \bibinfo{year}{2017}\natexlab{}.
\newblock \showarticletitle{SketchParse : Towards Rich Descriptions for Poorly
  Drawn Sketches using Multi-Task Hierarchical Deep Networks}. In
  \bibinfo{booktitle}{\emph{Proceedings of the 25th ACM International
  Conference on Multimedia}}. \bibinfo{pages}{10--18}.
\newblock


\bibitem[\protect\citeauthoryear{{Scarselli}, {Gori}, {Tsoi}, {Hagenbuchner},
  and {Monfardini}}{{Scarselli} et~al\mbox{.}}{2009}]%
        {GNN2009}
\bibfield{author}{\bibinfo{person}{F. {Scarselli}}, \bibinfo{person}{M.
  {Gori}}, \bibinfo{person}{A.~C. {Tsoi}}, \bibinfo{person}{M. {Hagenbuchner}},
  {and} \bibinfo{person}{G. {Monfardini}}.} \bibinfo{year}{2009}\natexlab{}.
\newblock \showarticletitle{The Graph Neural Network Model}.
\newblock \bibinfo{journal}{\emph{IEEE Transactions on Neural Networks}}
  \bibinfo{volume}{20}, \bibinfo{number}{1} (\bibinfo{year}{2009}),
  \bibinfo{pages}{61--80}.
\newblock


\bibitem[\protect\citeauthoryear{Schneider and Tuytelaars}{Schneider and
  Tuytelaars}{2016}]%
        {Schneider:2016:ESS}
\bibfield{author}{\bibinfo{person}{Ros\'{a}lia~G. Schneider} {and}
  \bibinfo{person}{Tinne Tuytelaars}.} \bibinfo{year}{2016}\natexlab{}.
\newblock \showarticletitle{Example-Based Sketch Segmentation and Labeling
  Using CRFs}.
\newblock \bibinfo{journal}{\emph{ACM Trans. Graph.}} \bibinfo{volume}{35},
  \bibinfo{number}{5} (\bibinfo{year}{2016}), \bibinfo{pages}{151:1--151:9}.
\newblock


\bibitem[\protect\citeauthoryear{Simonovsky and Komodakis}{Simonovsky and
  Komodakis}{2017}]%
        {Martin:2017:DEDF}
\bibfield{author}{\bibinfo{person}{Martin Simonovsky} {and}
  \bibinfo{person}{Nikos Komodakis}.} \bibinfo{year}{2017}\natexlab{}.
\newblock \showarticletitle{Dynamic Edge-Conditioned Filters in Convolutional
  Neural Networks on Graphs}. In \bibinfo{booktitle}{\emph{2017 IEEE Conference
  on Computer Vision and Pattern Recognition (CVPR)}}. \bibinfo{pages}{29--38}.
\newblock


\bibitem[\protect\citeauthoryear{{Song}, {Pang}, {Song}, {Xiang}, and
  {Hospedales}}{{Song} et~al\mbox{.}}{2018}]%
        {Song:2018:LSSCC}
\bibfield{author}{\bibinfo{person}{Jifei {Song}}, \bibinfo{person}{Kaiyue
  {Pang}}, \bibinfo{person}{Yi-Zhe {Song}}, \bibinfo{person}{Tao {Xiang}},
  {and} \bibinfo{person}{Timothy~M. {Hospedales}}.}
  \bibinfo{year}{2018}\natexlab{}.
\newblock \showarticletitle{Learning to Sketch with Shortcut Cycle
  Consistency}. In \bibinfo{booktitle}{\emph{2018 IEEE Conference on Computer
  Vision and Pattern Recognition (CVPR)}}. \bibinfo{pages}{801--810}.
\newblock


\bibitem[\protect\citeauthoryear{Tang and Liu}{Tang and Liu}{2009}]%
        {Tang:2009:RLLSD}
\bibfield{author}{\bibinfo{person}{Lei Tang} {and} \bibinfo{person}{Huan Liu}.}
  \bibinfo{year}{2009}\natexlab{}.
\newblock \showarticletitle{Relational Learning via Latent Social Dimensions}.
  In \bibinfo{booktitle}{\emph{Proceedings of the 15th ACM SIGKDD International
  Conference on Knowledge Discovery and Data Mining}}.
  \bibinfo{pages}{817--826}.
\newblock


\bibitem[\protect\citeauthoryear{Valsesia, Fracastoro, and Magli}{Valsesia
  et~al\mbox{.}}{2019}]%
        {Valsesia:2019:LLGM}
\bibfield{author}{\bibinfo{person}{Diego Valsesia}, \bibinfo{person}{Giulia
  Fracastoro}, {and} \bibinfo{person}{Enrico Magli}.}
  \bibinfo{year}{2019}\natexlab{}.
\newblock \showarticletitle{Learning Localized Generative Models for 3D Point
  Clouds via Graph Convolution}. In \bibinfo{booktitle}{\emph{International
  Conference on Learning Representations}}.
\newblock


\bibitem[\protect\citeauthoryear{Wang, Sun, Liu, Sarma, Bronstein, and
  Solomon}{Wang et~al\mbox{.}}{2019}]%
        {Wang:2019:DGCNN}
\bibfield{author}{\bibinfo{person}{Yue Wang}, \bibinfo{person}{Yongbin Sun},
  \bibinfo{person}{Ziwei Liu}, \bibinfo{person}{Sanjay~E. Sarma},
  \bibinfo{person}{Michael~M. Bronstein}, {and} \bibinfo{person}{Justin~M.
  Solomon}.} \bibinfo{year}{2019}\natexlab{}.
\newblock \showarticletitle{Dynamic Graph CNN for Learning on Point Clouds}.
\newblock \bibinfo{journal}{\emph{ACM Trans. Graph.}} \bibinfo{volume}{38},
  \bibinfo{number}{5} (\bibinfo{year}{2019}), \bibinfo{pages}{146:1--146:12}.
\newblock


\bibitem[\protect\citeauthoryear{Wertheimer}{Wertheimer}{1938}]%
        {Wertheimer:1938:Laws}
\bibfield{author}{\bibinfo{person}{Max Wertheimer}.}
  \bibinfo{year}{1938}\natexlab{}.
\newblock \showarticletitle{Laws of Organization in Perceptual Forms}.
\newblock \bibinfo{journal}{\emph{Psychologische Forschung}}
  \bibinfo{volume}{4} (\bibinfo{year}{1938}), \bibinfo{pages}{71--88}.
\newblock


\bibitem[\protect\citeauthoryear{Wu, Qi, Liu, and Yang}{Wu
  et~al\mbox{.}}{2018}]%
        {Wu:2018:SSeg}
\bibfield{author}{\bibinfo{person}{Xingyuan Wu}, \bibinfo{person}{Yonggang Qi},
  \bibinfo{person}{Jun Liu}, {and} \bibinfo{person}{Jie Yang}.}
  \bibinfo{year}{2018}\natexlab{}.
\newblock \showarticletitle{Sketchsegnet: A Rnn Model for Labeling Sketch
  Strokes}. In \bibinfo{booktitle}{\emph{2018 IEEE 28th International Workshop
  on Machine Learning for Signal Processing (MLSP)}}. \bibinfo{pages}{1--6}.
\newblock


\bibitem[\protect\citeauthoryear{Wu, Pan, Chen, Long, Zhang, and Yu}{Wu
  et~al\mbox{.}}{2019}]%
        {gcn_survey2019}
\bibfield{author}{\bibinfo{person}{Zonghan Wu}, \bibinfo{person}{Shirui Pan},
  \bibinfo{person}{Fengwen Chen}, \bibinfo{person}{Guodong Long},
  \bibinfo{person}{Chengqi Zhang}, {and} \bibinfo{person}{Philip~S. Yu}.}
  \bibinfo{year}{2019}\natexlab{}.
\newblock \showarticletitle{A Comprehensive Survey on Graph Neural Networks}.
\newblock \bibinfo{journal}{\emph{CoRR}}  \bibinfo{volume}{abs/1901.00596}
  (\bibinfo{year}{2019}).
\newblock
\showeprint[arxiv]{1901.00596}
\urldef\tempurl%
\url{http://arxiv.org/abs/1901.00596}
\showURL{%
\tempurl}


\bibitem[\protect\citeauthoryear{Xie, Xu, Mitra, Cohen-Or, Gong, Su, and
  Chen}{Xie et~al\mbox{.}}{2013}]%
        {sketch2design2013}
\bibfield{author}{\bibinfo{person}{Xiaohua Xie}, \bibinfo{person}{Kai Xu},
  \bibinfo{person}{Niloy Mitra}, \bibinfo{person}{Daniel Cohen-Or},
  \bibinfo{person}{Wenyong Gong}, \bibinfo{person}{Qi Su}, {and}
  \bibinfo{person}{Baoquan Chen}.} \bibinfo{year}{2013}\natexlab{}.
\newblock \showarticletitle{Sketch-to-Design: Context-Based Part Assembly}.
\newblock \bibinfo{journal}{\emph{Computer Graphics Forum}}
  \bibinfo{volume}{32} (\bibinfo{date}{11} \bibinfo{year}{2013}),
  \bibinfo{pages}{233--245}.
\newblock
\urldef\tempurl%
\url{https://doi.org/10.1111/cgf.12200}
\showDOI{\tempurl}


\bibitem[\protect\citeauthoryear{Xu, Hu, Leskovec, and Jegelka}{Xu
  et~al\mbox{.}}{2018a}]%
        {xu2018powerful}
\bibfield{author}{\bibinfo{person}{Keyulu Xu}, \bibinfo{person}{Weihua Hu},
  \bibinfo{person}{Jure Leskovec}, {and} \bibinfo{person}{Stefanie Jegelka}.}
  \bibinfo{year}{2018}\natexlab{a}.
\newblock \showarticletitle{How powerful are graph neural networks?}
\newblock \bibinfo{journal}{\emph{arXiv preprint arXiv:1810.00826}}
  (\bibinfo{year}{2018}).
\newblock


\bibitem[\protect\citeauthoryear{Xu, Kang, Fu, Sun, and Hu}{Xu
  et~al\mbox{.}}{2013}]%
        {Xu2013Sketch2Scene}
\bibfield{author}{\bibinfo{person}{Kun Xu}, \bibinfo{person}{Chen Kang},
  \bibinfo{person}{Hongbo Fu}, \bibinfo{person}{Wei-Lun Sun}, {and}
  \bibinfo{person}{Shi-Min Hu}.} \bibinfo{year}{2013}\natexlab{}.
\newblock \showarticletitle{Sketch2Scene: Sketch-based Co-retrieval and
  Co-placement of 3D Models}.
\newblock \bibinfo{journal}{\emph{Acm Transactions on Graphics (Proceedings
  SIGGRAPH)}} \bibinfo{volume}{32}, \bibinfo{number}{4} (\bibinfo{year}{2013}),
  \bibinfo{pages}{1--15}.
\newblock


\bibitem[\protect\citeauthoryear{Xu, Li, Tian, Sonobe, Kawarabayashi, and
  Jegelka}{Xu et~al\mbox{.}}{2018b}]%
        {2018Representation}
\bibfield{author}{\bibinfo{person}{Keyulu Xu}, \bibinfo{person}{Chengtao Li},
  \bibinfo{person}{Yonglong Tian}, \bibinfo{person}{Tomohiro Sonobe},
  \bibinfo{person}{Ken~Ichi Kawarabayashi}, {and} \bibinfo{person}{Stefanie
  Jegelka}.} \bibinfo{year}{2018}\natexlab{b}.
\newblock \showarticletitle{Representation Learning on Graphs with Jumping
  Knowledge Networks}.
\newblock \bibinfo{journal}{\emph{arXiv}} (\bibinfo{year}{2018}).
\newblock


\bibitem[\protect\citeauthoryear{Yi, Kim, Ceylan, Shen, Yan, Su, Lu, Huang,
  Sheffer, and Guibas}{Yi et~al\mbox{.}}{2016}]%
        {2016ASAF}
\bibfield{author}{\bibinfo{person}{Li Yi}, \bibinfo{person}{Vladimir~G. Kim},
  \bibinfo{person}{Duygu Ceylan}, \bibinfo{person}{I~Chao Shen},
  \bibinfo{person}{Mengyan Yan}, \bibinfo{person}{Hao Su},
  \bibinfo{person}{Cewu Lu}, \bibinfo{person}{Qixing Huang},
  \bibinfo{person}{Alla Sheffer}, {and} \bibinfo{person}{Leonidas Guibas}.}
  \bibinfo{year}{2016}\natexlab{}.
\newblock \showarticletitle{A scalable active framework for region annotation
  in 3D shape collections}.
\newblock \bibinfo{journal}{\emph{ACM Transactions on Graphics (TOG)}}
  \bibinfo{volume}{35}, \bibinfo{number}{6cd} (\bibinfo{year}{2016}),
  \bibinfo{pages}{210.1--210.12}.
\newblock


\bibitem[\protect\citeauthoryear{Ying, He, Chen, Eksombatchai, Hamilton, and
  Leskovec}{Ying et~al\mbox{.}}{2018}]%
        {Ying:2018:GCN}
\bibfield{author}{\bibinfo{person}{Rex Ying}, \bibinfo{person}{Ruining He},
  \bibinfo{person}{Kaifeng Chen}, \bibinfo{person}{Pong Eksombatchai},
  \bibinfo{person}{William~L. Hamilton}, {and} \bibinfo{person}{Jure
  Leskovec}.} \bibinfo{year}{2018}\natexlab{}.
\newblock \showarticletitle{Graph Convolutional Neural Networks for Web-Scale
  Recommender Systems}. In \bibinfo{booktitle}{\emph{Proceedings of the 24th
  ACM SIGKDD International Conference on Knowledge Discovery and Data Mining}}.
  \bibinfo{pages}{974--983}.
\newblock


\bibitem[\protect\citeauthoryear{Zhu, Xiao, and Zheng}{Zhu
  et~al\mbox{.}}{2019}]%
        {zhu20192d}
\bibfield{author}{\bibinfo{person}{Xianyi Zhu}, \bibinfo{person}{Yi Xiao},
  {and} \bibinfo{person}{Yan Zheng}.} \bibinfo{year}{2019}\natexlab{}.
\newblock \showarticletitle{2D freehand sketch labeling using CNN and CRF}.
\newblock \bibinfo{journal}{\emph{Multimedia Tools and Applications}}
  (\bibinfo{year}{2019}), \bibinfo{pages}{1--18}.
\newblock


\end{thebibliography}

\end{document}